\documentclass{article}

\usepackage[final]{neurips_2025}

\usepackage[utf8]{inputenc} 
\usepackage[T1]{fontenc}    
\usepackage{hyperref}       
\usepackage{url}            
\usepackage{booktabs}       
\usepackage{amsfonts}       
\usepackage{nicefrac}       
\usepackage{microtype}      
\usepackage{xcolor}         
\usepackage{graphicx}
\usepackage{lipsum}
\usepackage{float}
\usepackage{wrapfig}
\usepackage{adjustbox}

\usepackage{algorithm}
\usepackage{algpseudocode}

\usepackage{graphicx}
\usepackage{amsmath}
\usepackage{amssymb}
\usepackage{booktabs}
\usepackage{multirow}
\usepackage{enumitem}
\usepackage[table]{xcolor}
\usepackage{amssymb,amsmath,amsthm,amsfonts,bm}
\usepackage{multirow}
\usepackage{booktabs}
\usepackage{tabularx}
\usepackage[table]{xcolor}
\usepackage{pgf}
\usepackage{cancel}
\usepackage{wrapfig}
\usepackage{microtype}
\usepackage{xspace}
\usepackage{pifont}
\usepackage{scalerel}
\usepackage{graphicx}
\usepackage{tikz}
\usepackage{mathtools}
\usepackage{svg}
\usepackage{hyperref}
\usepackage{dsfont}
\usepackage{enumitem}

\usepackage[dvipsnames]{xcolor}

\definecolor{cliptta_color}{rgb}{1, 0.97, 0.92}
\definecolor{lightgray}{rgb}{0.95, 0.95, 0.95}
\definecolor{BrickRed}{rgb}{0.8, 0.25, 0.33}
\definecolor{OliveGreen}{rgb}{0.33, 0.42, 0.18}
\definecolor{DarkGreen}{RGB}{0,100,0}
\definecolor{DarkRed}{RGB}{139,0,0}

\newcommand{\better}[1]{\textcolor{DarkGreen}{\scriptsize{\,\,($\uparrow$#1)}}}
\newcommand{\worse}[1]{\textcolor{DarkRed}{\scriptsize{\,\,\,($\downarrow$#1)}}}
\newcommand{\same}[1]{\phantom{\scriptsize{\,\,($\downarrow$#1)}}}

\newcommand{\mypar}[1]{\vspace{3pt}\noindent\textbf{#1~}}

\newcolumntype{Y}{>{\centering\arraybackslash}X}

\newcommand{\ie}{\emph{i.e., }}

\newcommand{\XX}{\mathbf{X}}

\newcommand{\xx}{\mathbf{x}}
\newcommand{\yy}{\mathbf{y}}

\DeclareMathOperator*{\argmax}{arg\,max}
\DeclareMathOperator*{\argmin}{arg\,min}

\theoremstyle{plain}

\theoremstyle{definition}

\theoremstyle{remark}

\newcommand{\ccol}{\cellcolor{orange!20}}
\definecolor{top2}{RGB}{255, 230, 160}   
\definecolor{top4}{RGB}{220, 240, 255}   
\definecolor{top6}{RGB}{210, 255, 210}   

\newcommand{\Real}{\mathbb{R}}

\newcommand{\AAA}{\mathbf{A}}
\newcommand{\BB}{\mathbf{B}}
\newcommand{\CC}{\mathbf{C}}
\newcommand{\KK}{\mathbf{K}}

\newcommand{\hh}{\mathbf{h}}

\usepackage{glossaries}

\newacronym{cnn}{CNN}{Convolutional Neural Network}
\newacronym{vit}{ViT}{Vision Transformer}
\newacronym{nlp}{NLP}{Natural Language Processing}
\newacronym{vlm}{VLM}{Vision-language Model}
\newacronym{ssm}{SSM}{State Space Model}
\newacronym{ss2d}{SS2D}{2D Selective Scan}
\newacronym{ood}{OOD}{out-of-distribution}
\newacronym{tta}{TTA}{Test Time Adaptation}
\newacronym{iid}{i.i.d}{Independently and Identically Distributed}
\newacronym{zoh}{ZOH}{zero-order hold}
\newacronym{bn}{BN}{Batch Normalization}
\newacronym{trust}{TRUST}{\textbf{T}est-Time \textbf{R}efinement using \textbf{U}ncertainty-Guided \textbf{S}SM \textbf{T}raverses}
\newacronym{rrm}{RRM}{Robust Risk Minimization}

\usepackage{xcolor}                       

\hypersetup{
    colorlinks=true,
    linkcolor=black,
    citecolor=black,
    filecolor=black,
    urlcolor=black
}

\title{TRUST: Test-Time Refinement using Uncertainty-Guided SSM Traverses}

\author{\normalfont Sahar Dastani$^{1,2}$\thanks{Equal contribution} \and 
Ali Bahri$^{1}$\footnotemark[1] \and 
Gustavo Adolf Vargas Hakim$^{1}$ \and 
Moslem Yazdanpanah$^{1}$ \and
Mehrdad Noori$^{1}$ \and   
David Osowiechi$^{1}$ \and  
Samuel Barbeau$^{1}$ \and  
Ismail Ben Ayed$^{1}$ \and  
Herve Lombaert$^{2,3}$ \and 
Christian Desrosiers$^{1}$\\
$^1$LIVIA, ILLS, ÉTS Montréal, Canada, \\ $^2$Mila - Quebec AI Institute, $^3$Polytechnique Montreal\\ 
}

\begin{document}

\maketitle

\begin{abstract}
State Space Models (SSMs) have emerged as efficient alternatives to Vision Transformers (ViTs), with VMamba standing out as a pioneering architecture designed for vision tasks. However, their generalization performance degrades significantly under distribution shifts. To address this limitation, we propose TRUST (Test-Time Refinement using Uncertainty-Guided SSM Traverses), a novel test-time adaptation (TTA) method that leverages diverse traversal permutations to generate multiple causal perspectives of the input image. Model predictions serve as pseudo-labels to guide updates of the Mamba-specific parameters, and the adapted weights are averaged to integrate the learned information across traversal scans. Altogether, TRUST is the first approach that explicitly leverages the unique architectural properties of SSMs for adaptation. Experiments on seven benchmarks show that TRUST consistently improves robustness and outperforms existing TTA methods. The
code is available at: \url{https://github.com/Sahardastani/trust}.
\end{abstract}    
\vspace{-10pt}
\section{Introduction}
\vspace{-5pt}

The field of visual representation learning has advanced rapidly due to the ability of deep neural networks to extract rich and generalizable features. \glspl{cnn} \cite{simonyan2014very, he2016deep, huang2017densely, tan2019efficientnet, liu2022convnet} excel in modeling local patterns through strong inductive biases but struggle with global context. \glspl{vit} \cite{dosovitskiy2020image, liu2021swin, zhang2023hivit, touvron2021training} address this challenge using a self-attention mechanism, although at higher computational cost. More recently, \glspl{ssm} \cite{gu2021efficiently, fu2022hungry, smith2022simplified} emerged as a scalable alternative, providing global receptive fields with linear complexity. Despite their efficiency, the performance of \glspl{ssm} degrades under distribution shift. This is mainly due to violations of the \gls{iid} assumption, which is often disrupted in real-world settings. While generalization strategies are well developed for \glspl{cnn} and \glspl{vit}, vision-specific strategies for \glspl{ssm} are still lacking.

This work focuses on \emph{VMamba} \cite{liu2024vmamba}, a vision-adapted variant of the Mamba architecture~\cite{gu2023mamba}, designed for sequential visual processing. VMamba introduces \gls{ss2d}, a four-way traversal mechanism that scans image patches along predefined spatial directions. However, this directional processing introduces a strong inductive bias by aligning internal representations with fixed traversal paths \cite{dastani2025spectral}, which may hinder generalization under distribution shifts. Additionally, the hidden states of VMamba store historical context over the traversal sequence. When exposed to unseen domains, this context may accumulate domain-specific artifacts, leading to amplified bias during propagation and ultimately degrading generalization~\cite{dgmamba}.

\begin{figure}[t]
    \centering
    \includegraphics[trim=0 680 0 0, clip, width=\textwidth]{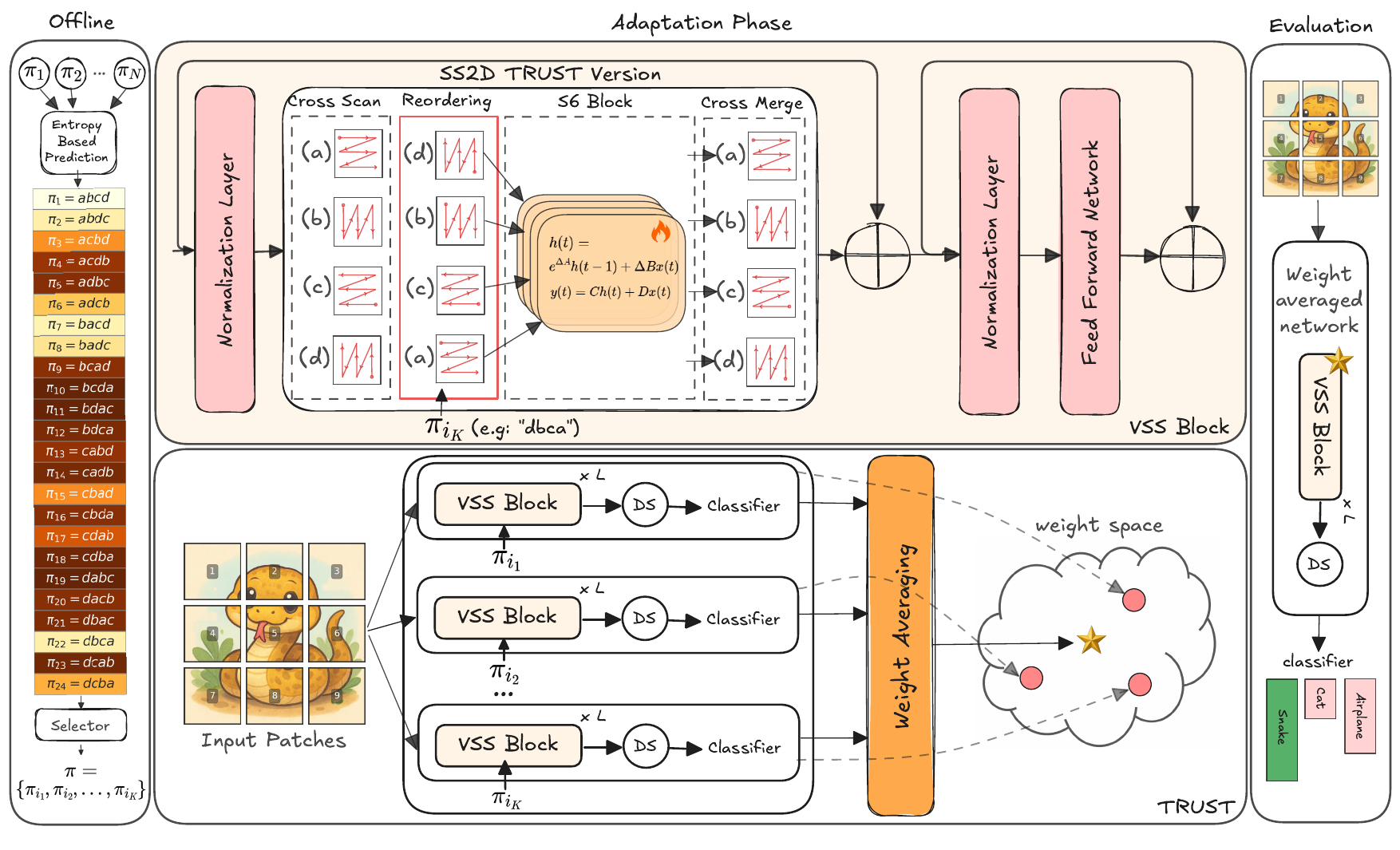}
    \vspace{-15pt}
    \caption{An overview of the proposed method. Our network consists of three stages: offline, adaptation, and evaluation. In the Offline stage, multiple traversal permutations are generated and ranked by entropy. The top-K most confident permutations are selected. During adaptation, each permutation reorders the cross-scan traversals, and the Mamba-specific state-space parameters are updated accordingly. The resulting models are then merged via parameter-space averaging. The final averaged model is used for inference in evaluation stage.}
    \label{main_figure}
\end{figure}

To address these challenges, we propose \gls{trust}, a novel \gls{tta} strategy specifically designed for \glspl{ssm} that leverages VMamba's internal traversal mechanism. As illustrated in the offline phase of Figure~\ref{main_figure}, different traversal permutations result in varying prediction entropy levels, revealing the sensitivity of the model to causal ordering. \gls{trust} systematically generates multiple traversal permutations by reordering the four directional scans, exposing the model to diverse causal perspectives of the same input. At test time, we compute the entropy of predictions from each permutation and select the ones with the lowest entropy, which are associated with more stable and domain-robust hidden states. 

Inspired by previous work on flat minima and weight averaging \cite{keskar2016large, swad}, our method aggregates the outputs of different traversal permutations to implicitly explore a broader and flatter region of the loss landscape. Each permutation activates a distinct computational pathway through the model, effectively sampling different local minima. Selecting the top-k traversal permutations with the lowest predictive entropy and computing a weighted average of their model parameters mitigates the impact of noisy or uncertain predictions. This enhances generalization without requiring source data or model retraining.

We outline the main contributions of our work as follows:
\begin{itemize}[leftmargin=2em]
    \item We propose \gls{trust}, the first \gls{tta} approach specifically designed for Mamba-based vision models. Unlike prior methods that depend on data augmentation or auxiliary models, our approach takes advantage of the internal traversal dynamics of VMamba to improve generalization under distribution shift.
    \item \gls{trust} introduces a novel weighted averaging strategy to promote robustness, which permutes traversal directions and averages model weights from the most confident paths. 
    \item We validate our method on \textbf{seven standard benchmarks}, where it consistently outperforms existing \gls{tta} methods, establishing new state-of-the-art results.
\end{itemize}

\vspace{-10pt}
\section{Related Work}
\vspace{-5pt}

\textbf{\acrfull{tta}} is a paradigm of Domain Adaptation (DA)~\cite{shot} that intends to enhance model generalization during deployment. Specifically, a pre-trained model is adapted to incoming data batches without requiring label supervision or having access to source data. As a pioneering approach, Tent~\cite{tent} minimizes prediction entropy as an adaptation objective, based on the principle that distribution shifts in the target domain reduce the confidence of a model in its predictions. Examples of Tent-based approaches include using confidence thresholds for sample selection before adaptation~\cite{eta,ostta}, minimizing the sharpness of the entropy loss~\cite{sar}, using class-balanced memory queues for better adaptation sampling~\cite{rotta}, and meta-learning the entropy loss~\cite{conjugate}, among other alternatives~\cite{sotta,stamp,unient}. 
The literature also includes several other adaptation strategies that do not minimize entropy, such as contrastive learning~\cite{cotta,adacontrast}, Laplacian optimization~\cite{lame,program}, and prototype-based pseudo-labeling~\cite{tast}. While flexible, these techniques do not capitalize on the unique properties of specific architectures such as VMamba. 

\textbf{Architecture-specific \gls{tta}}. Recent advances in \gls{tta} have introduced a variety of model-specific strategies. Among those, FOA~\cite{foa} loosely adheres to ViTs by incorporating gradient-free learning of additional prompts in the form of token embeddings. For \glspl{vlm}~\cite{tpt,tda,watt,clipartt}, recent strategies incorporate text information as an auxiliary tool for adaptation. Examples include prompt learning via entropy minimization~\cite{tpt}, positive and negative caching with text-based pseudo-labels~\cite{tda}, transductive adaptation using conformal pseudo-captions~\cite{clipartt}, and weight averaging across text templates~\cite{watt}. SSM-oriented \gls{tta} has also been explored to a smaller extent. In a recent work, STAD~\cite{stad} uses Markov processes to learn time-varying prototypes for classifying examples during inference and adapts the classification head of the model.

\textbf{Weight Averaging for Robust Adaptation.} Initially introduced in the context of Domain Generalization (DG)~\cite{metadg,mixstyle,fds}, weight averaging has gained popularity as an effective technique for enhancing model robustness to domain shifts~\cite{swa,swad}. Notably, methods like SWAD~\cite{swad} aim to identify flat minima in the loss landscape by minimizing loss fluctuations across diverse inputs. This is often accomplished by averaging the weights of models trained on different image augmentations. While similar ideas have been adopted in the \gls{tta} literature~\cite{sar,watt,sampling, purge_gate, smart_pc}, they typically rely on explicit input diversification, which introduces additional computational overhead. In contrast, our approach leverages the intrinsic diversity of SSM-based traversals to induce variation directly in the weights of our model, without requiring external data augmentations.

\vspace{-10pt}
\section{Method}
\vspace{-5pt}

Section~\ref{preliminary} introduces key concepts in \gls{tta} and \gls{ssm}, laying the groundwork for our approach. In Section~\ref{traverse}, we propose traversal permutations to generate complementary causal perspectives of the input, followed by a weight averaging strategy in Section~\ref{wa} for robust and efficient inference. Figure~\ref{main_figure} shows an overview of our method.

\subsection{Preliminaries}\label{preliminary}

\mypar{\acrfull{tta} in Image Classification.} Consider a classification model composed of a feature processor $f_\theta$, parameterized by $\theta$, and a classifier $h_\phi$ parameterized by $\phi$. The model is pre-trained on a source dataset $\mathcal{D}_s = \{(\XX_j, y_j)\}_{j=1}^{N_s}$ consisting of images $\XX_j \in \Real^{W \times H \times 3}$ and corresponding labels $y_j \in \{1,\ldots,C\}$, where $C$ is the number of classes. Formally, the model learns a map $F_s: \mathcal{X}_s \rightarrow \mathcal{Y}_s$ between the input distribution $\mathcal{X}_s$ and output distribution $\mathcal{Y}_s$ from the source. Training on $\mathcal{D}_s$ is performed using a supervised loss $\mathcal{L}_{\text{train}}$ (typically cross-entropy). At test-time, the model is exposed to a target dataset $\mathcal{D}_t = \{\XX_j\}_{j=N_s+1}^{N_s+N_t}$, which lacks labels and follows a different distribution, giving rise to the map $F_t: \mathcal{X}_t \rightarrow \mathcal{Y}_t$. Although the source and target datasets share the same label space ($\mathcal{Y}_s = \mathcal{Y}_t$), their joint distribution differ, \ie $P(\mathcal{X}_s,\mathcal{Y}_s) \neq P(\mathcal{X}_t, \mathcal{Y}_t)$. Such difference is referred to as \emph{covariate domain shift}~\cite{lame}.

\mypar{\acrfull{ssm} Formulation.} \gls{ssm}-based models \cite{gu2021efficiently, gu2023mamba} map a 1D input sequence $\xx(t) \in \Real$ to an output $\yy(t) \in \Real$ through a learnable hidden state $\hh(t) \in \Real^N$, governed by a linear dynamical system:
\begin{equation}
    \hh'(t) = \AAA \hh(t) + \BB \xx(t), \quad \yy(t) = \CC \hh(t), \label{mamba}
\end{equation}
where $\AAA \in \mathbb{R}^{N \times N}$, $\BB \in \mathbb{R}^{N \times 1}$, and $\CC \in \mathbb{R}^{1 \times N}$ are learnable parameters. To integrate this formulation into deep learning, it is discretized using the zero-order hold (ZOH) method with step size $\Delta \in \mathbb{R}$, resulting in the discrete recurrence:
\begin{equation}
    \hh(t) = \overline{\AAA} \hh(t-1) + \overline{\BB} \xx(t), \quad \yy(t) = \CC \hh(t),  \quad \texttt{where} \quad \overline{\AAA} = \exp(\Delta \AAA), \quad \overline{\BB} \approx \Delta \BB.
\end{equation}
Unrolling this recurrence over time yields a 1D convolution with kernel $\overline{\KK} \in \mathbb{R}^L$, where $L \in \mathbb{N}$ denotes the kernel length. This form enables efficient parallel computation during training.
\begin{equation}
    \yy = \xx \odot \overline{\KK}, \quad \overline{\KK} = \left( \CC \overline{\BB}, \CC \overline{\AAA} \overline{\BB}, \ldots, \CC \overline{\AAA}^{L-1} \overline{\BB} \right).
\end{equation}
\mypar{SS2D for 2D Visual Processing.} To extend 1D state-space models to visual data, VMamba introduces the \gls{ss2d} module, which reshapes an input image $\XX \in \mathbb{R}^{H \times W \times C}$ into a sequence of $T$ non-overlapping patches $\{\xx_t\}_{t=1}^{T}$. As shown in Fig. \ref{main_figure}, these patches are traversed in a predefined causal order determined by one of four canonical directions using the Cross-Scan module:  left-to-right ($a$), top-to-bottom ($b$),  right-to-left ($c$), and bottom-to-top ($d$). For each direction, the patch sequence is processed recurrently by a shared, discretized 1D \gls{ssm}, ensuring causality and directional consistency. The outputs from all four scans are then aggregated using the Cross-Merge module.

\subsection{Traversal Permutation}\label{traverse}


To address the limitations of VMamba under distribution shifts, we propose a \gls{tta} strategy that leverages VMamba's intrinsic directional traversal mechanism. We define a set of traversal permutations \begin{equation} 
\mathcal{P} = \{\pi_1, \pi_2 \dots, \pi_N\},
\end{equation} 
where each $\pi_i$ represents a unique ordering of the four canonical scan directions, with $\pi_1 = \{a, b, c, d\}$ corresponding to the original ordering used in VMamba (see Fig. \ref{main_figure} for the definition of directions $a$, $b$, $c$ and $d$). We assess the predictive confidence of the model under each traversal permutation $\pi_i \in \mathcal{P}$ by computing the \textit{Shannon entropy} of its output distribution. Permutations are then ranked in ascending order of entropy, and the top-$K$ permutations with the lowest entropy are selected (the offline mode of Figure~\ref{main_figure}): 
\begin{equation} 
\mathcal{P}_\texttt{selected} = \{\pi_{i_1}, \pi_{i_2}, \dots, \pi_{i_K} \}. 
\end{equation} 

During the adaptation phase, for each selected permutation $\pi_{i_k}$, $k=1,\ldots,K$, the input is processed according to the corresponding traversal order. Denoting $p(\XX; \, \pi_{i_k}) \in [0,1]^C$ as the output of the model for an image $\XX$ when using traversal permutation $\pi_{i_k}$, we compute a pseudo-label $\hat{y}_{k}$ by taking the class with maximum predicted probability: 
\begin{equation}
\hat{y}_{k}  = \argmax_{c \in \{1,\ldots,C\}} \, \big[p(\XX; \, \pi_{i_k})\big]_c. 
\end{equation}
The Mamba-specific parameters are then updated to minimize the average cross-entropy (\emph{log loss}) between the model prediction and the corresponding pseudo-label over each target sample in batch $\mathcal{B}$:
\begin{equation} 
\boldsymbol{\theta}_k = \argmin_{\boldsymbol{\theta}} \, 
- \frac{1}{|\mathcal{B}|} \sum_{\XX \in \mathcal{B}} \log \left[p(\XX; \, \pi_{i_k})\right]_{\hat{y}_{k}}.
\end{equation}
This is achieved using back-propagation as in standard methods. After optimization, the adapted parameter set $\boldsymbol{\theta}_k$ is cached, allowing a subsequent ensemble-based aggregation at inference.

Our method processes the same image through multiple directional permutations, enabling VMamba to exploit complementary causal views of the input. These distinct trajectories expose the model to both global consistency (identical token set) and local variation (different hidden‑state evolutions), which helps find a flatter minima offering a stronger \gls{ood} generalization \cite{swad}. Crucially, re‑ordering prevents domain‑specific artifacts from always entering the recurrence at the same time step. 
If an artifact $\boldsymbol{\varepsilon}$ is embedded in a corrupted patch $\xx_{t_{\boldsymbol{\varepsilon}}}$, then under the default traversal $\pi_1$, this patch always appears at time step $t = t_{\boldsymbol{\varepsilon}}$ in the processing sequence. The hidden state update at that step becomes:
\begin{equation}
    \hh^{(1)}(t_{\boldsymbol{\varepsilon}}) = f\left(\hh^{(1)}(t_{\boldsymbol{\varepsilon}} - 1),\, \xx_{t_{\boldsymbol{\varepsilon}}} + \boldsymbol{\varepsilon} \right).
\end{equation}

Since the hidden state $\hh(t)$ is recurrent and accumulates over time, injecting $\boldsymbol{\varepsilon}$ early in the sequence allows its influence to propagate through many subsequent updates. Therefore, the impact of the artifact depends heavily on the position of $t_{\boldsymbol{\varepsilon}}$ within the sequence. In our method, we apply a traversal permutation $\pi_{i_k}$, which changes the order in which patches are processed. Under $\pi_{i_k}$, the corrupted patch $\xx_{t_{\boldsymbol{\varepsilon}}}$ appears at time step $t = \pi_{i_k}(t_{\boldsymbol{\varepsilon}})$, leading to the updated hidden state:
\begin{equation}
    \hh^{(i)}\left(\pi_{i_k}(t_{\boldsymbol{\varepsilon}})\right) = f\left(\hh^{(i)}\left(\pi_{i_k}(t_{\boldsymbol{\varepsilon}}) - 1\right),\, \xx_{t_{\boldsymbol{\varepsilon}}} + \boldsymbol{\varepsilon} \right).
\end{equation}

Across different permutations, the corrupted patch enters the sequence at varying time steps. This shifts the effect of artifact to different hidden states, breaking its consistent influence pattern. During test-time adaptation, these variations allow the model to learn diverse responses, and the subsequent weight averaging of adapted models helps suppress artifact-induced bias.

\subsection{Leveraging Traversal Permutations Through Weight Averaging}\label{wa}

To improve both stability and generalization under distribution shifts, we aggregate the adapted models obtained from the top-$K$ traversal permutations. During the evaluation phase, we utilize the averaged weights 
\begin{equation}
\bar{\theta} = \frac{1}{K} \sum_{k=1}^{K} \theta_k
\end{equation}
and evaluate the model using the default traversal path $\pi_1$. In this setting, each $\theta_k$ represents an adapted model under a distinct traversal permutation. This results in a single, consolidated model that captures the benefits of multi-directional adaptation. 

This simple yet effective averaging strategy draws inspiration from the notion of \textit{flat minima} in the loss landscape \cite{swa,swad}, where parameter configurations that lie in flat regions tend to exhibit greater generalization and robustness to perturbations. Weight averaging encourages convergence to a region in parameter space with low curvature, thereby reducing sensitivity to input variations and improving performance under distribution shifts. In our approach, the integration of traversal permutations with weight averaging is inspired by \gls{rrm} \cite{swad}. \gls{rrm} aims to minimize the worst-case empirical loss within a neighborhood around the current parameters, formulated as:
\begin{equation}
\hat{\mathcal{E}}_{\mathcal{D}}^{\gamma}(\theta) \, = \, \max_{\|\Delta\| \leq \gamma} \, \hat{\mathcal{E}}_{\mathcal{D}}(\theta + \Delta),
\end{equation}

\begin{wrapfigure}{r}{0.45\textwidth}
  \vspace{-10pt}
  \centering
  \includegraphics[width=0.9\linewidth]{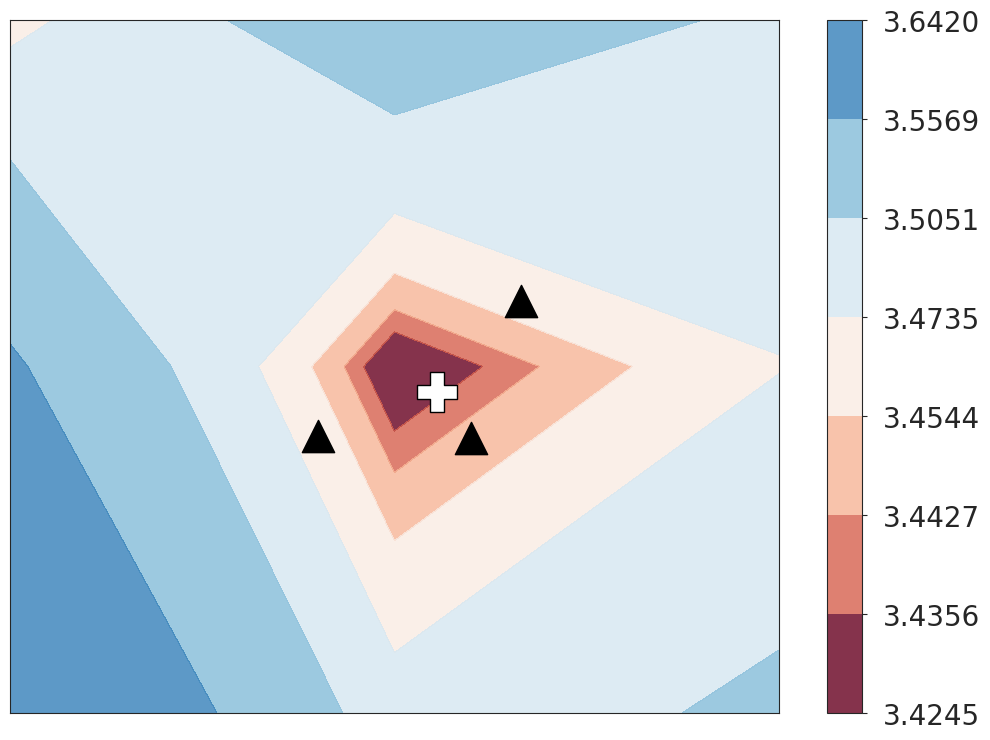}
  \vspace{-10pt}
  \caption{Loss surface of model parameters.}
  \label{loss}
\end{wrapfigure}

where $\gamma > 0$ defines the neighborhood around the model parameters $\theta$, and $\Delta$ shows small perturbations. To further illustrate this point, Figure~\ref{loss} represents the test loss surface over model parameters under Gaussian noise corruption from the ImageNet-C dataset. Each triangle marks a model adapted via a different traversal permutation, while the central cross denotes the weight-averaged model. The axes depict linear interpolations between parameter sets, offering a 2D view of the high-dimensional landscape. This visualization demonstrates how different traversal permutations lead to diverse optima, and how weight averaging converges toward a smoother, lower-loss region, enhancing robustness and generalization.

Unlike traditional \gls{tta} approaches that rely on data augmentation, external source data, or auxiliary objectives to generate diverse model variants for weight averaging, our approach leverages the internal architecture of VMamba, specifically, its canonical traversal mechanism. By permuting traversal scans, we induce structural variability that yields diverse perspectives of the same input, enabling effective adaptation without external augmentation.

\vspace{-10pt}
\section{Experiments}
\vspace{-5pt}

In this section, we present a comprehensive evaluation of our proposed method across seven benchmark datasets. We begin by describing the datasets used for evaluation, followed by implementation details and the baselines employed. We then report our main results and conclude with a set of ablation studies to analyze key components of our approach.

\subsection{Datasets}
We evaluate the performance and generalization of our method across a wide range of \gls{tta} benchmarks, covering corruption robustness and domain generalization. For corruption-based robustness, we use CIFAR10-C \cite{corruption}, CIFAR100-C \cite{corruption}, and ImageNet-C \cite{corruption}, which apply 15 corruption types (e.g., noise, blur, weather, digital artifacts) at five severity levels. These datasets scale from 10 (CIFAR10-C) to 100 (CIFAR100-C) to 1000 (ImageNet-C) classes, enabling systematic analysis of model degradation.

For domain generalization, we assess on PACS \cite{pacs}, ImageNet-S \cite{sketch}, ImageNet-V2 \cite{v2}, and ImageNet-R \cite{R}. PACS includes four visual styles with seven shared classes, using leave-one-domain-out evaluation. ImageNet-S features sketch-style abstractions; ImageNet-V2 provides a cleaner, independently collected test set; and ImageNet-R introduces style shifts across 200 classes with diverse artistic renditions. Together, these benchmarks test adaptability to unseen distributions, styles, and abstractions.

\subsection{Implementation Details}
We perform adaptation exclusively on the Mamba-specific state space parameters within the \gls{ss2d} module of each VMamba block, while keeping the rest of the model frozen. The base VMamba model is trained with batch normalization layers. The adaptation is guided using a pseudo-labeling strategy, where confident model predictions are treated as supervision targets in a cross-entropy loss. This enables self-supervised adaptation without requiring access to ground-truth labels. The adaptation proceeds in an online manner, with model updates applied sequentially (i.e., without resetting the weights to their pre-adaptation values). Optimization is performed using the Adam optimizer with a learning rate of $10^{-4}$ and a batch size of $128$, ensuring consistent dynamics and fair comparison across benchmarks. All experiments were conducted using a single NVIDIA $A6000$ GPU.

\subsection{Baselines}
We compare our method against several state-of-the-art \gls{tta} approaches. Source Only refers to the performance of VMamba without any adaptation. ETA \cite{eta} minimizes an entropy loss modulated by a sample-adaptive weighting mechanism, updating only the affine parameters of normalization layers. LAME \cite{lame} enforces smooth decision boundaries by introducing Laplacian regularization over the model predictions. SAR \cite{sar} promotes robustness by explicitly optimizing for flat minima using sharpness-aware minimization. SHOT  \cite{shot} aligns the classifier by minimizing entropy while encouraging confident and diverse predictions, using a fixed feature extractor. Tent \cite{tent} adapts the model by updating only the affine parameters of normalization layers via entropy minimization.


We evaluate two variants of our approach: \gls{trust} naive, which adapts only the Mamba-specific \gls{ss2d} parameters, and \gls{trust}, which further enhances robustness by averaging over multiple traversal permutations during adaptation.

\begin{table*}[htbp]
\setlength\tabcolsep{4pt}
\centering
\resizebox{0.9\textwidth}{!}{
\small
\begin{tabular}{l|l|ccccccccccccccc|c}
\toprule
& Method & \rotatebox{90}{gaussian noise} & \rotatebox{90}{shot noise} & \rotatebox{90}{impulse noise} & \rotatebox{90}{defocus blur} & \rotatebox{90}{glass blur} & \rotatebox{90}{motion blur} & \rotatebox{90}{zoom blur} & \rotatebox{90}{frost} & \rotatebox{90}{snow} & \rotatebox{90}{fog} & \rotatebox{90}{brightness} & \rotatebox{90}{contrast} & \rotatebox{90}{elastic} & \rotatebox{90}{pixelate} & \rotatebox{90}{jpeg compression} & Mean \\
\midrule
\parbox[t]{4mm}{\multirow{7}{*}{\rotatebox[origin=c]{90}{CIFAR10-C}}} 
& Source only           & 46.8 & 48.4 & 45.0 & 73.5 & 52.6 & 73.0 & 78.7 & 71.8 & 75.8 & 77.3 & 85.7 & 69.6 & 63.7 & 67.9 & 59.0 & 65.9\same{0.0} \\
& ETA \cite{eta}        & 46.7 & 48.3 & 44.8 & 73.5 & 52.6 & 73.0 & 78.6 & 75.7 & 71.4 & 77.2 & 85.7 & 69.6 & 63.7 & 67.9 & 59.0 & 65.8\worse{0.1} \\
& LAME \cite{lame}      & 46.7 & 48.3 & 44.8 & 73.5 & 52.6 & 73.0 & 78.6 & 71.8 & 75.8 & 77.2 & 85.7 & 69.6 & 63.7 & 67.9 & 59.0 & 65.9\same{0.0}\\
& SAR \cite{sar}        & 47.7 & 49.5 & 46.2 & 74.3 & 53.4 & 73.8 & 79.1 & 72.5 & 76.5 & 78.0 & 86.1 & 70.8 & 64.5 & 68.9 & 60.0 & 66.8\better{0.9}\\
& SHOT \cite{shot}      & 47.8 & 49.7 & 46.3 & 74.3 & 53.7 & 74.0 & 79.3 & 72.6 & 76.6 & 78.1 & 86.3 & 70.7 & 64.5 & 68.9 & 59.9 & 66.8\better{0.9} \\
& Tent \cite{tent}      & 47.3 & 49.2 & 45.8 & 74.2 & 53.1 & 73.7 & 79.1 & 72.2 & 76.3 & 77.9 & 86.1 & 70.4 & 64.3 & 68.7 & 59.6 & 66.5\better{0.6} \\
& \gls{trust} naive            & 58.9 & 61.8 & 62.0 & 79.8 & 60.9 & 79.1 & 82.6 & 80.5 & 81.8 & 83.6 & 88.8 & 81.8 & 70.3 & 75.1 & 66.0 & 74.2\better{8.3} \\
& \ccol \gls{trust}            & \ccol \textbf{63.1} & \ccol \textbf{67.8} & \ccol \textbf{70.3} & \ccol \textbf{81.0} & \ccol \textbf{64.5} & \ccol \textbf{81.4} & \ccol \textbf{85.0} & \ccol \textbf{83.2} & \ccol \textbf{85.4} & \ccol \textbf{85.8} & \ccol \textbf{90.1} & \ccol \textbf{85.7} & \ccol \textbf{72.1} & \ccol \textbf{79.1} & \ccol \textbf{68.6} & \ccol \textbf{77.5\better{11.6}} \\
\midrule
\parbox[t]{4mm}{\multirow{7}{*}{\rotatebox[origin=c]{90}{CIFAR100-C}}} 
& Source only           & 21.0 & 22.1 & 18.3 & 50.6 & 27.7 & 51.0 & 56.2 & 45.3 & 50.6 & 52.4 & 65.3 & 43.2 & 39.0 & 41.7 & 33.4 & 41.2\same{0.0} \\
& ETA \cite{eta}        & 21.2 & 22.3 & 18.7 & 50.8 & 27.8 & 51.2 & 56.3 & 45.4 & 50.8 & 52.7 & 65.5 & 43.5 & 39.2 & 42.0 & 33.7 & 41.4\better{0.2} \\
& LAME \cite{lame}      & 21.0 & 22.1 & 18.3 & 50.6 & 27.7 & 51.0 & 56.2 & 45.3 & 50.6 & 52.5 & 65.4 & 43.2 & 39.0 & 41.7 & 33.4 & 41.2\same{0.0} \\
& SAR \cite{sar}        & 21.9 & 22.8 & 19.3 & 51.1 & 28.2 & 51.5 & 56.7 & 46.4 & 51.4 & 53.1 & 65.8 & 44.2 & 39.9 & 42.8 & 34.2 & 41.9\better{0.7} \\
& SHOT \cite{shot}      & 21.9 & 22.9 & 19.1 & 51.4 & 28.3 & 51.8 & 56.8 & 46.3 & 51.4 & 53.3 & 66.0 & 44.2 & 39.8 & 42.7 & 34.3 & 42.0\better{0.8} \\
& Tent \cite{tent}      & 21.6 & 22.6 & 18.9 & 51.1 & 28.2 & 51.5 & 56.7 & 46.2 & 51.1 & 53.1 & 65.8 & 44.0 & 39.7 & 42.5 & 34.0 & 41.8\better{0.6} \\
& \gls{trust} naive            & 32.1 & 32.8 & 34.1 & 56.9 & 35.4 & 57.2 & 61.6 & 54.6 & 57.8 & 60.1 & 69.6 & 55.6 & 46.6 & 50.8 & 41.0 & 49.8\better{8.6} \\
& \ccol \gls{trust}            & \ccol \textbf{37.8} & \ccol \textbf{38.9} & \ccol \textbf{42.3} & \ccol \textbf{60.9} & \ccol \textbf{36.6} & \ccol \textbf{60.8} & \ccol \textbf{65.4} & \ccol \textbf{59.0} & \ccol \textbf{62.2} & \ccol \textbf{64.5} & \ccol \textbf{71.7} & \ccol \textbf{63.1} & \ccol \textbf{50.3} & \ccol \textbf{56.6} & \ccol \textbf{44.9} & \ccol \textbf{54.3\better{13.1}} \\
\midrule
\parbox[t]{4mm}{\multirow{7}{*}{\rotatebox[origin=c]{90}{ImageNet-C}}} 
& Source only           & 24.3 & 26.1 & 25.1 & 22.2 & 23.2 & 35.4 & 43.2 & 49.3 & 48.4 & 56.9 & 70.0 & 26.8 & 45.1 & 43.7 & 41.4 & 38.7\same{0.0} \\
& ETA \cite{eta}        & 26.4 & 28.4 & 27.2 & 23.5 & 24.6 & 37.2 & 45.1 & 50.8 & 51.0 & 58.8 & 70.6 & 29.1 & 47.7 & 46.9 & 45.0 & 40.8\better{2.1} \\
& LAME \cite{lame}      & 24.3 & 26.1 & 25.1 & 22.2 & 23.2 & 35.4 & 43.2 & 49.3 & 48.4 & 56.9 & 70.0 & 26.8 & 45.1 & 43.7 & 41.4 & 38.8\better{0.1} \\
& SAR \cite{sar}        & 26.5 & 29.2 & 28.0 & 24.5 & 25.3 & 37.4 & 45.1 & 51.0 & 51.7 & 59.1 & 70.5 & 31.5 & 48.2 & 48.6 & 46.3 & 41.5\better{2.8} \\
& SHOT \cite{shot}      & 28.0 & 30.1 & 28.8 & 25.0 & 26.0 & 38.0 & 45.7 & 51.0 & 51.5 & 59.1 & 70.6 & 30.2 & 48.4 & 47.8 & 45.8 & 41.7\better{3.0} \\
& Tent \cite{tent}      & 27.8 & 30.0 & 28.8 & 24.9 & 25.9 & 38.0 & 45.5 & 51.0 & 51.3 & 59.1 & 70.6 & 30.0 & 48.2 & 47.8 & 45.7 & 41.7\better{3.0} \\
& \gls{trust} naive            & 43.4 & 45.6 & 44.9 & 38.3 & 36.6 & 53.0 & 54.9 & 57.1 & 60.2 & 66.0 & 72.2 & 50.2 & 59.0 & 61.1 & 58.5 & 53.4\better{14.7} \\
& \ccol \gls{trust}            & \ccol \textbf{46.8} & \ccol \textbf{49.4} & \ccol \textbf{48.5} & \ccol \textbf{42.8} & \ccol \textbf{40.8} & \ccol \textbf{57.1} & \ccol \textbf{57.9} & \ccol \textbf{57.3} & \ccol \textbf{61.7} & \ccol \textbf{66.8} & \ccol \textbf{71.9} & \ccol \textbf{54.9} & \ccol \textbf{61.4} & \ccol \textbf{63.6} & \ccol \textbf{60.2} & \ccol \textbf{56.1\better{17.4}} \\
\bottomrule
\end{tabular}
}
\caption{Top-1 classification accuracy (\%) under various corruption types on CIFAR10-C, CIFAR100-C, and ImageNet-C datasets. Increases/decreases in mean accuracy compared to performing no adaptation (Source only) is highlighted in green/red color.}
\label{main_online}
\end{table*}

\subsection{Main Results}
Table~\ref{main_online} presents the Top-1 accuracy under the highest corruption severity (level 5) for CIFAR10-C, CIFAR100-C, and ImageNet-C datasets. \gls{trust} consistently outperforms all baselines across the three datasets, demonstrating strong generalization under severe distribution shifts. It also surpasses its naive variant, underscoring the effectiveness of permutation-based strategy in enhancing robustness.

\mypar{CIFAR10-C.} \gls{trust} achieves a mean accuracy of $77.5\%$, outperforming both Tent and SHOT by margins of $11.0\%$ and $10.7\%$, respectively. On challenging corruptions, \gls{trust} yields the largest improvements over SHOT, such as elastic ($7.5\%$), glass blur ($10.8\%$), and motion blur ($7.4\%$). Compared to its naive variant, \gls{trust} provides a further $3.3\%$ boost. These results highlight the benefits of traversal diversity in enhancing corruption robustness.

\mypar{CIFAR100-C.} On a more fine-grained benchmark, \gls{trust} attains a mean accuracy of $54.3\%$, exceeding SHOT and SAR by $12.3\%$ and $12.4\%$, respectively. The largest absolute improvements over SHOT are observed under challenging corruptions such as impulse noise ($23.2\%$), contrast ($18.9\%$), and shot noise ($16.1\%$), demonstrating the robustness of our method under severe distribution shifts. It also improves upon \gls{trust} naive by $4.5\%$.

\mypar{ImageNet-C.} \gls{trust} achieves a strong mean accuracy of $56.1\%$ on the large-scale ImageNet-C benchmark, outperforming both Tent and SHOT by $14.4\%$, and \gls{trust} naive by $2.7\%$. On challenging corruptions, \gls{trust} yields the largest improvements over SHOT, such as glass blur ($14.8\%$), elastic ($19.3\%$), and jpeg compression ($14.4\%$), reflecting the advantage of our method in mitigating both spatial distortions and digital artifacts.

In addition to corruption-specific robustness, our method demonstrates strong generalization to broader distribution shifts, as summarized in Table~\ref{domain_shift_online}. On \textbf{ImageNet-S}, our method achieves $41.5\%$ accuracy, outperforming SHOT by a substantial margin of $8.9\%$, and providing a modest $0.4\%$ improvement over the naive variant. For \textbf{ImageNet-V2}, \gls{trust} attains $64.0\%$, exceeding SHOT by $1.6\%$ and providing a $0.6\%$ gain over its naive variant. The most significant boost is observed on the \textbf{ImageNet-R} benchmark, which features real-world renditions of ImageNet categories. Here, our method achieves $44.3\%$, surpassing both Tent and SHOT by $12.4\%$, and outperforming \gls{trust} naive by $4.6\%$. On the \textbf{PACS} dataset, our model reaches $69.9\%$ accuracy, offering consistent improvements over SHOT and Tent by $2.5\%$, and over the naive variant by $2.8\%$. These consistent gains across varied benchmarks highlight the adaptability of our permutation-based strategy, which not only improves corruption robustness but also scales effectively to domain generalization tasks.

\begin{table*}[htbp]
\setlength\tabcolsep{4pt}
\centering
\resizebox{0.9\textwidth}{!}{
\small
\begin{tabular}{l|ccccccc}
\toprule
\textbf{Method} & CIFAR10-C & CIFAR100-C & ImageNet-C & ImageNet-S & ImageNet-V2 & ImageNet-R & PACS \\
\midrule
Source only           & 65.9\same{0.0} & 41.2\same{0.0} & 38.7\same{0.0} & 31.4\same{0.0} & 62.2\same{0.0} & 31.3\same{0.0} & 66.7\same{0.0} \\
ETA \cite{eta}        & 65.8\worse{0.1} & 41.4\better{0.2} & 40.8\better{2.1} & 31.4\same{0.0} & 62.2\same{0.0} & 31.3\same{0.0} & 66.7\same{0.0} \\
LAME \cite{lame}      & 65.9\same{0.0} & 41.2\same{0.0} & 38.8\better{0.1} & 31.4\same{0.0} & 62.2\same{0.0} & 31.3\same{0.0} & 66.7\same{0.0} \\
SAR \cite{sar}        & 66.8\better{0.9} & 41.9\better{0.7} & 41.5\better{2.8} & 32.6\better{1.2} & 62.4\better{0.2} & 32.0\better{0.7} & 67.3\better{0.6} \\
SHOT \cite{shot}      & 66.8\better{0.9} & 42.0\better{0.8} & 41.7\better{3.0} & 32.6\better{1.2} & 62.4\better{0.2} & 31.9\better{0.6} & 67.4\better{0.7} \\
Tent \cite{tent}      & 66.5\better{0.6} & 41.8\better{0.6} & 41.7\better{3.0} & 32.5\better{1.1} & 62.3\better{0.1} & 31.9\better{0.6} & 67.4\better{0.7} \\                    
\gls{trust} naive            & 74.2\better{8.3} & 49.8\better{8.6} & 53.4\better{14.7} & 41.1\better{9.7} & 63.4\better{1.2} & 39.7\better{8.4} & 67.1\better{0.4} \\
\ccol \gls{trust}            & \ccol \textbf{77.5\better{11.6}} & \ccol \textbf{54.3\better{13.1}} & \ccol \textbf{56.1\better{17.4}} & \ccol \textbf{41.5\better{10.1}} & \ccol \textbf{64.0\better{1.8}} & \ccol \textbf{44.3\better{13.0}} & \ccol \textbf{69.9\better{3.2}} \\
\bottomrule
\end{tabular}
}
\caption{Top-1 classification accuracy (\%) across datasets. For CIFAR10-C, CIFAR100-C, and ImageNet-C, values are averaged over all corruptions; for ImageNet-S, V2, R, and PACS, they reflect test set accuracy. Increases/decreases in mean accuracy compared to performing no adaptation (Source only) is highlighted in green/red color.}
\label{domain_shift_online}
\end{table*}

\subsection{Ablation Study}
In this section, we present a comprehensive ablation study on the CIFAR10-C dataset to assess the impact of key factors on the performance of our model. Specifically, we analyze the effects of batch size, augmentation types, number of traversal permutations, number of adaptation iterations, aggregation strategies, effect of traversing type in evaluation and computational overhead.

\begin{figure}[h]
  \centering
  \begin{minipage}[h]{0.48\textwidth}
    \includegraphics[width=\linewidth]{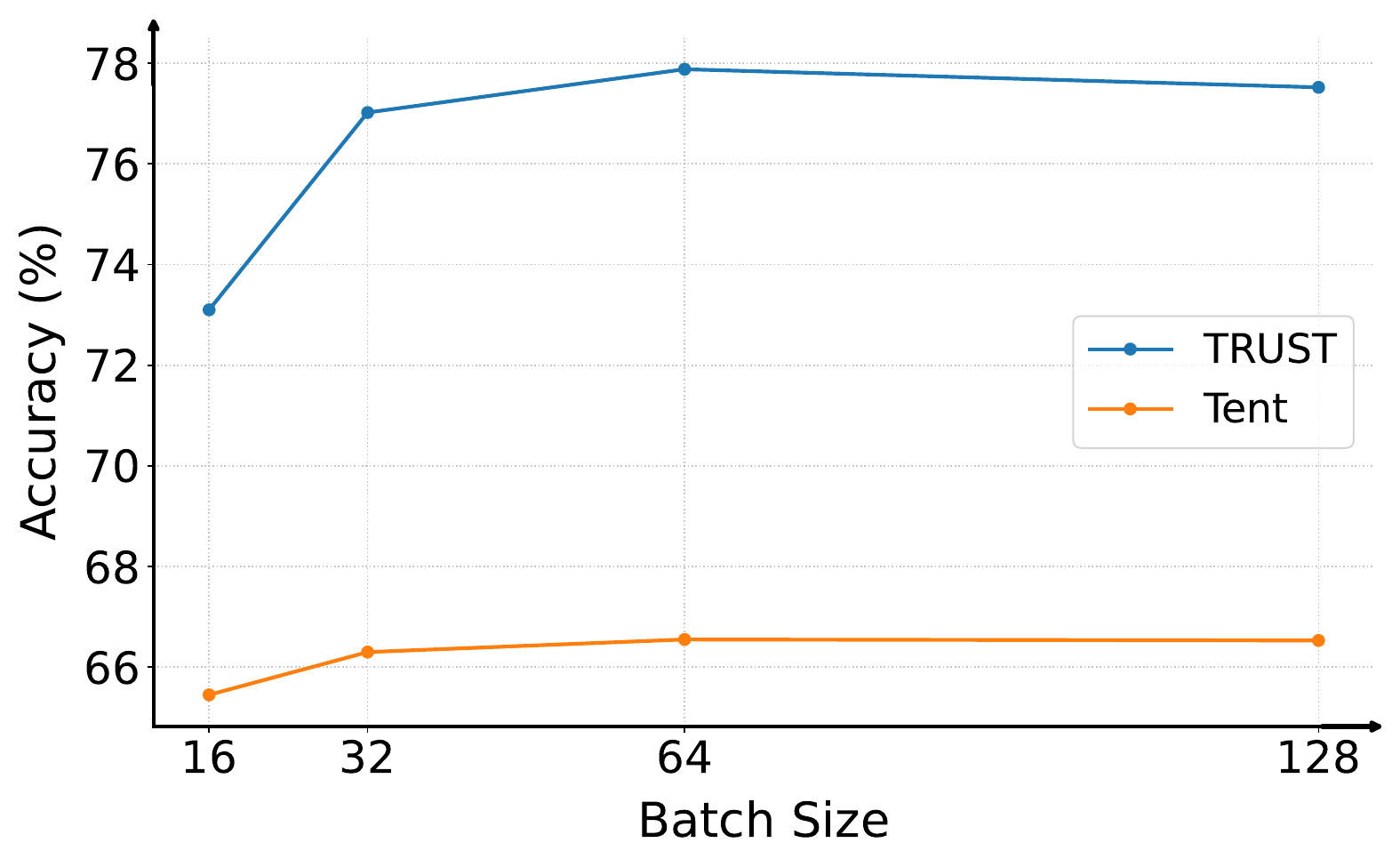}
    \vspace{-10pt}
    \caption{Accuracy comparison between \gls{trust} and Tent across varying batch sizes on CIFAR10-C dataset.}
    \label{batch_size}
  \end{minipage}\hfill
  \begin{minipage}[h]{0.48\textwidth}
    \includegraphics[width=\linewidth]{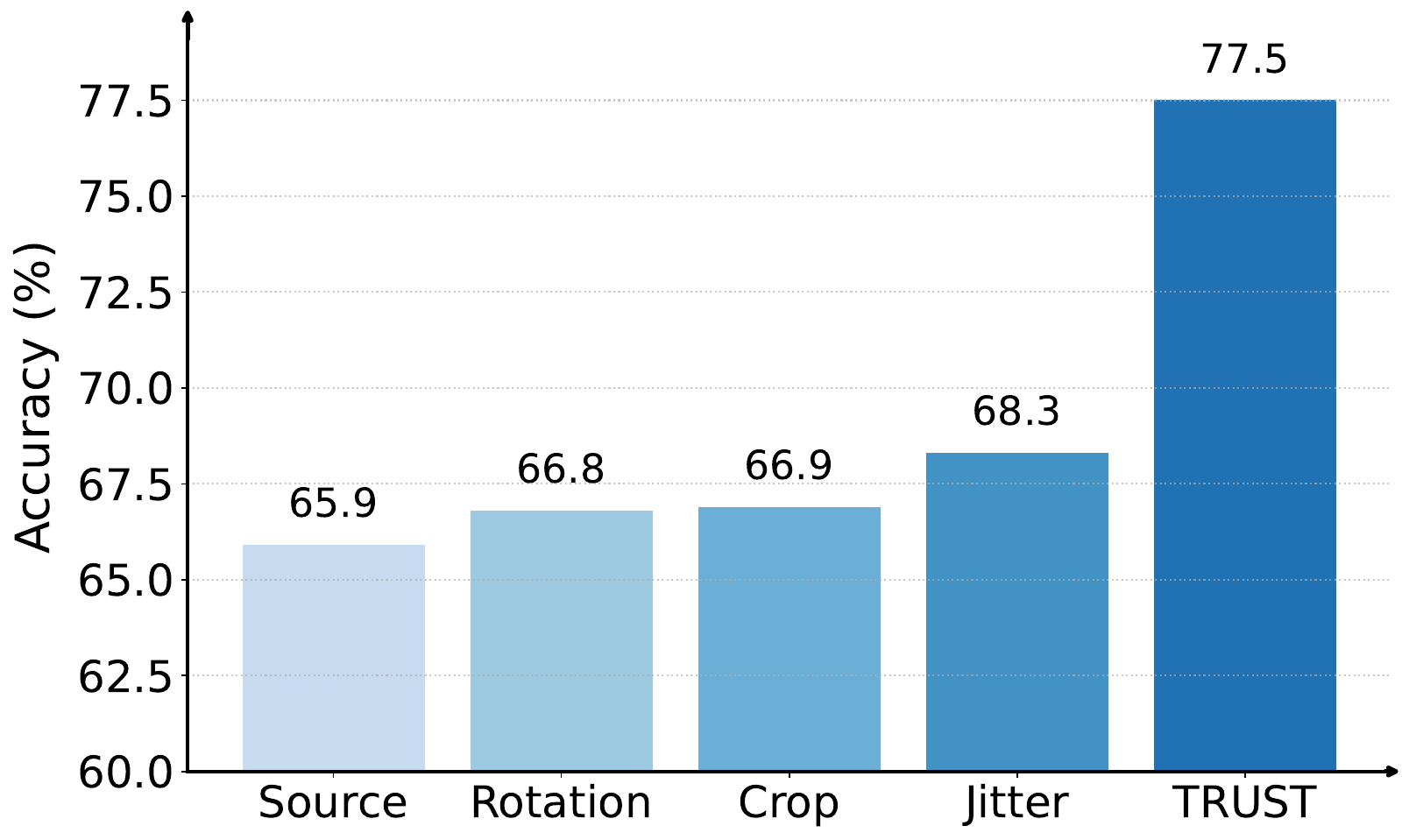}
    \vspace{-10pt}
    \caption{Performance comparison between standard augmentations and \gls{trust} on CIFAR10-C dataset.}
    \label{augmentation}
  \end{minipage}
\end{figure}

\mypar{Batch Size.} Figure~\ref{batch_size} demonstrates that \gls{trust} maintains a consistent advantage over Tent across different batch sizes. Even at smaller sizes such as 16 and 32, it delivers an accuracy improvement of around $8\%$. This indicates that our approach remains reliable and effective, even when operating with limited data per batch.

\mypar{Augmentation Impact.}
To evaluate whether standard data augmentations can serve as effective alternatives to our traversal permutation strategy, we compare \gls{trust} against common augmentations including rotation, random cropping, and color jitter. As shown in Figure~\ref{augmentation}, these traditional augmentations yield only modest improvements over the baseline source-only model on CIFAR10-C, with accuracies ranging from 65.9\% to 68.3\%. In contrast, our method achieves a substantial performance boost, reaching 77.5\%, which is significantly higher than all augmentation baselines. This indicates that simple augmentations fail to sufficiently address distribution shifts.

\begin{figure}[h]
  \centering
  \begin{minipage}[h]{0.48\textwidth}
    \includegraphics[width=\linewidth]{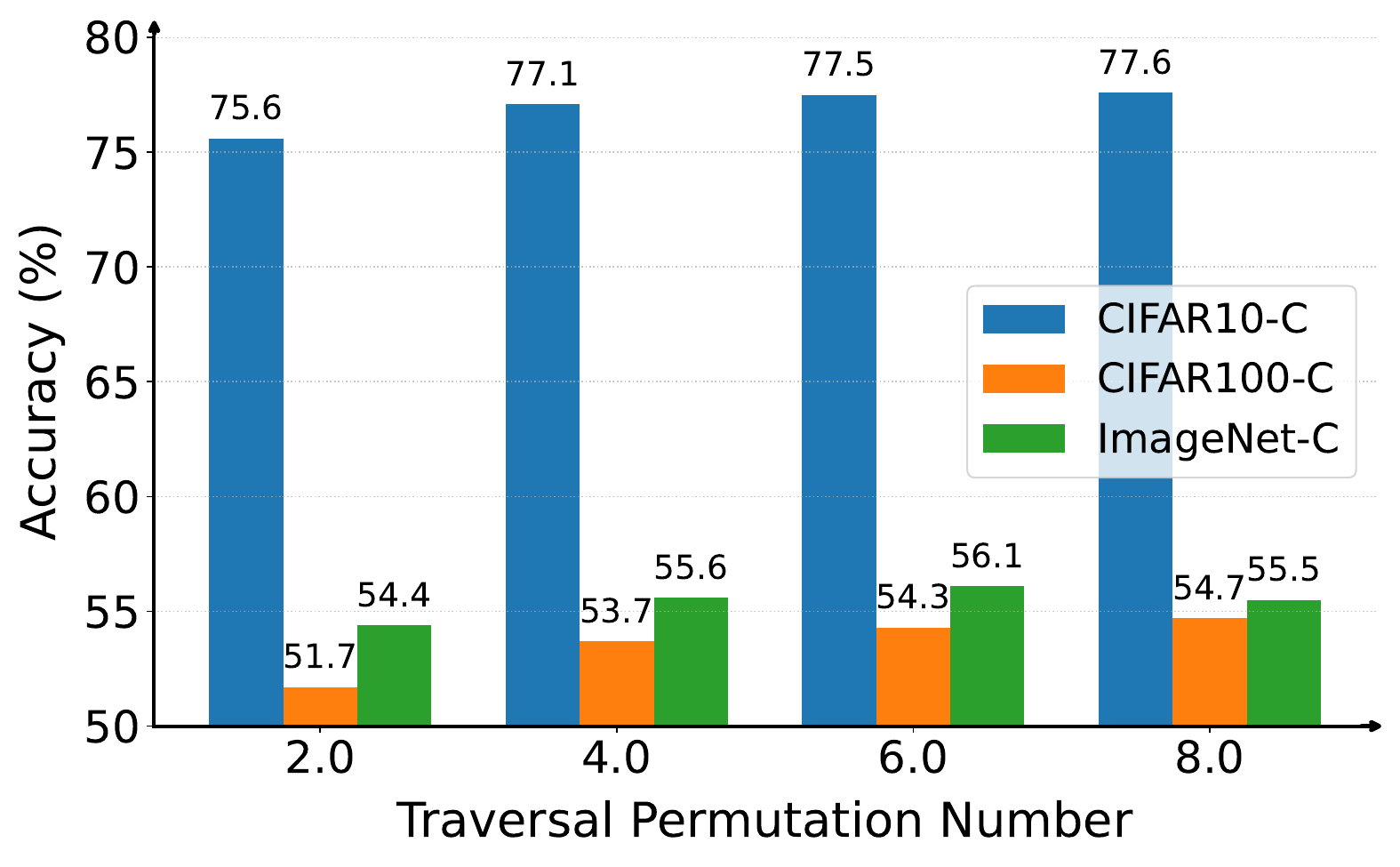}
    \vspace{-10pt}
    \caption{Effect of traversal permutation count on accuracy across three datasets.}
    \label{variation}
  \end{minipage}\hfill
  \begin{minipage}[h]{0.48\textwidth}
    \includegraphics[width=\linewidth]{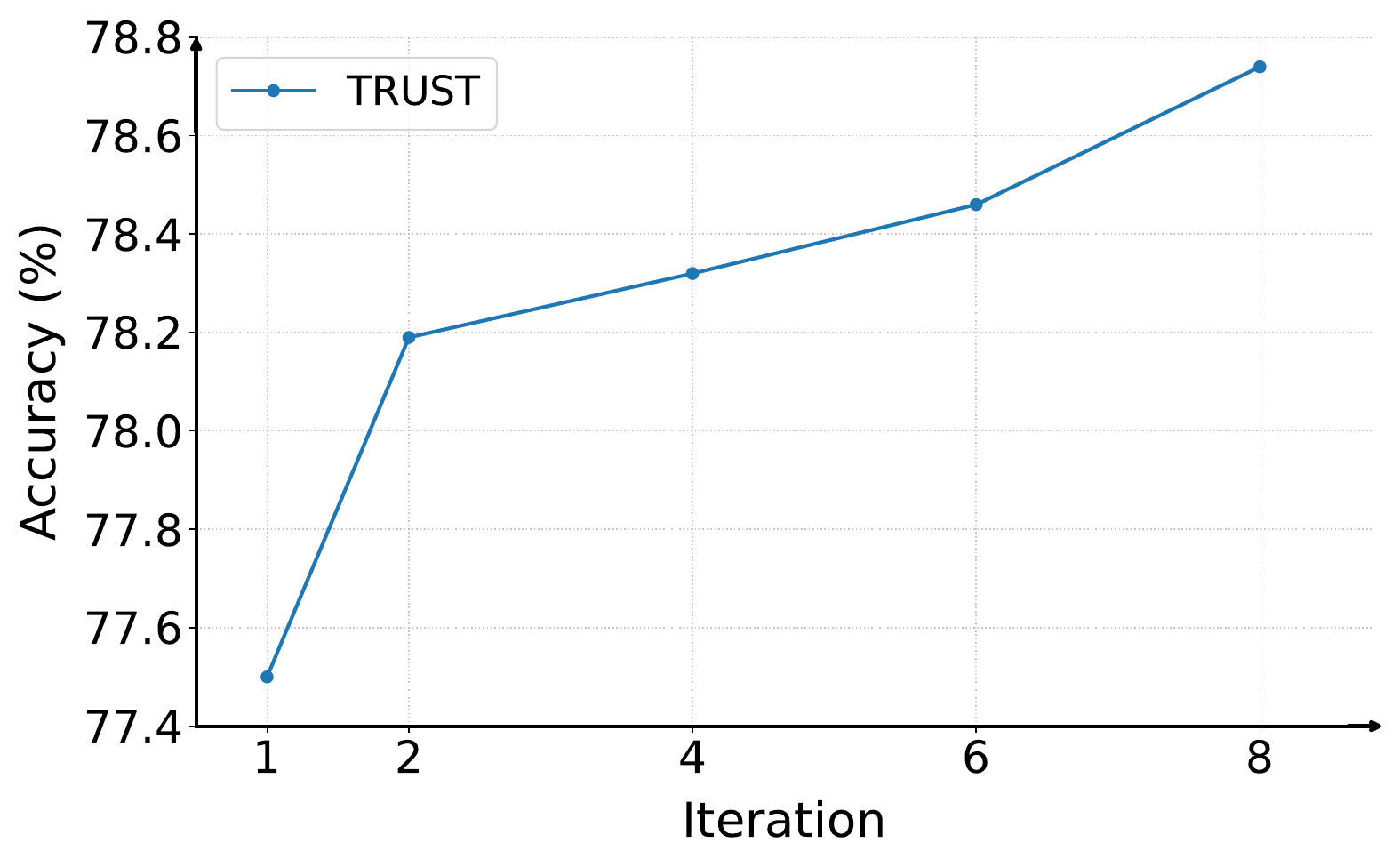}
    \vspace{-10pt}
    \caption{Model performance across adaptation iterations on CIFAR10-C dataset.}
    \label{iteration}
  \end{minipage}
\end{figure}

\mypar{Number of Traversal Permutations.} Figure~\ref{variation} shows how accuracy varies with the number of traversal permutations across CIFAR10-C, CIFAR100-C, and ImageNet-C. Overall, increasing the number of permutations consistently improves performance on all datasets. This is because by applying multiple traversal permutations, we expose the model to diverse causal perspectives of the same input. This variation encourages the model to learn different adaptation patterns at test time. Increasing the number of traversal permutations introduces more variation and, as shown in works such as Model-Soups \cite{model_soups}, a greater number of diverse, correctly-aligned representations (our top-K permutations) can enhance the effectiveness of weight averaging.

More specifically on CIFAR10-C, accuracy rises from 75.6\% (2 permutations) to 77.6\% (8 permutations). For CIFAR100-C, the gain is from 51.7\% to 54.7\%, and for ImageNet-C, from 54.4\% to a peak of 56.1\% at six permutations before slightly dropping to 55.5\% at eight. These results confirm that incorporating multiple traversal permutations enhances robustness under distribution shifts. Although accuracy improves with more permutations, gains diminish after six, suggesting that six permutations strike a good balance between performance and computational efficiency. For a breakdown of memory overhead, see Figure~\ref{cc}.

\mypar{Number of Adaptation Iterations.} Figure~\ref{iteration} illustrates that accuracy improves progressively with an increasing number of iterations. Despite the potential for further improvement with more iterations, we chose to use only a single iteration in subsequent experiments to enable faster adaptation without compromising on competitive performance.

\begin{figure}[h]
  \centering
  \begin{minipage}[h]{0.48\textwidth}
    \includegraphics[width=\linewidth]{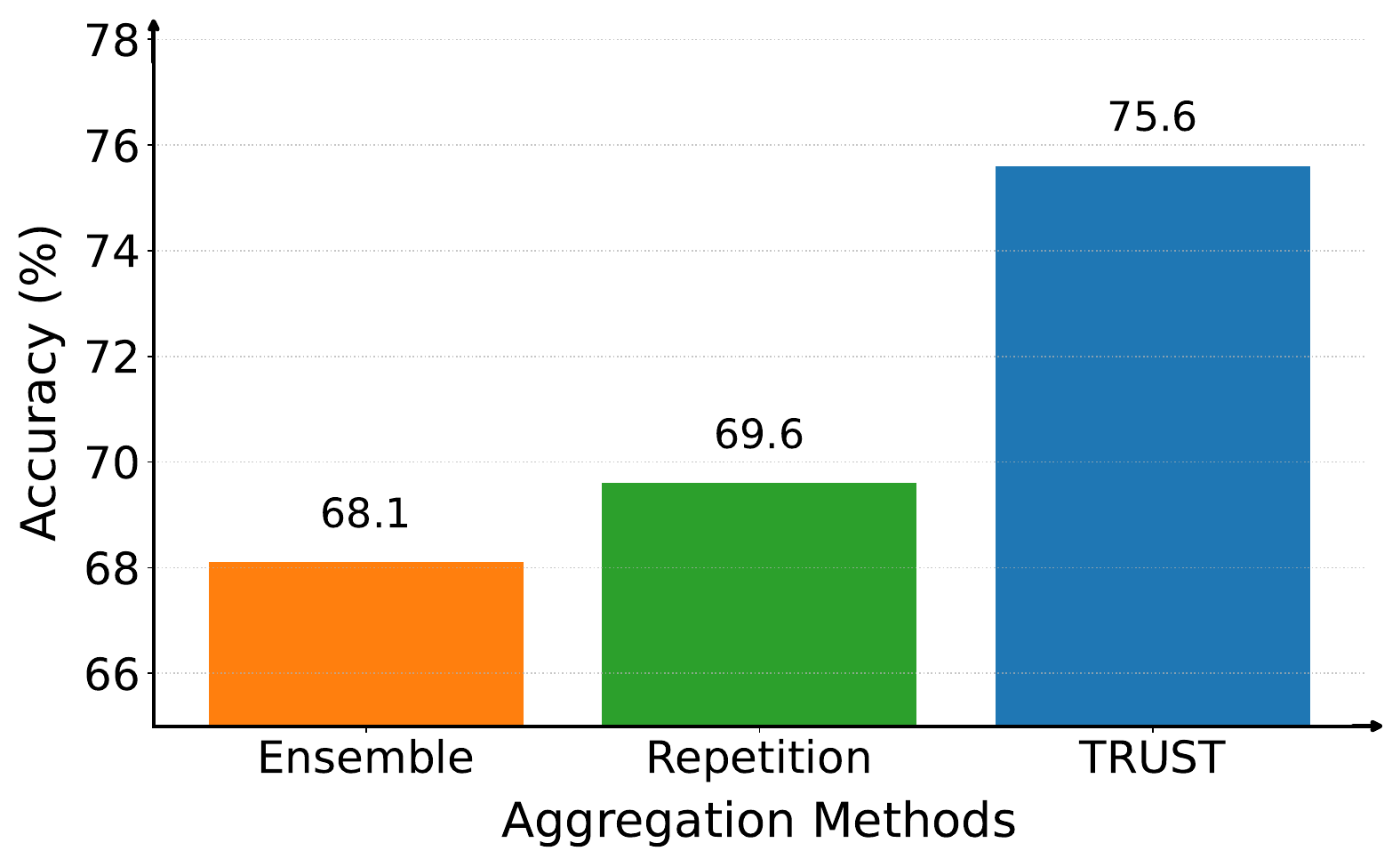}
    \vspace{-10pt}
    \caption{Accuracy comparison of different aggregation strategies on CIFAR10-C dataset.}
    \label{aggregation}
  \end{minipage}\hfill
  \begin{minipage}[h]{0.48\textwidth}
    \includegraphics[width=\linewidth]{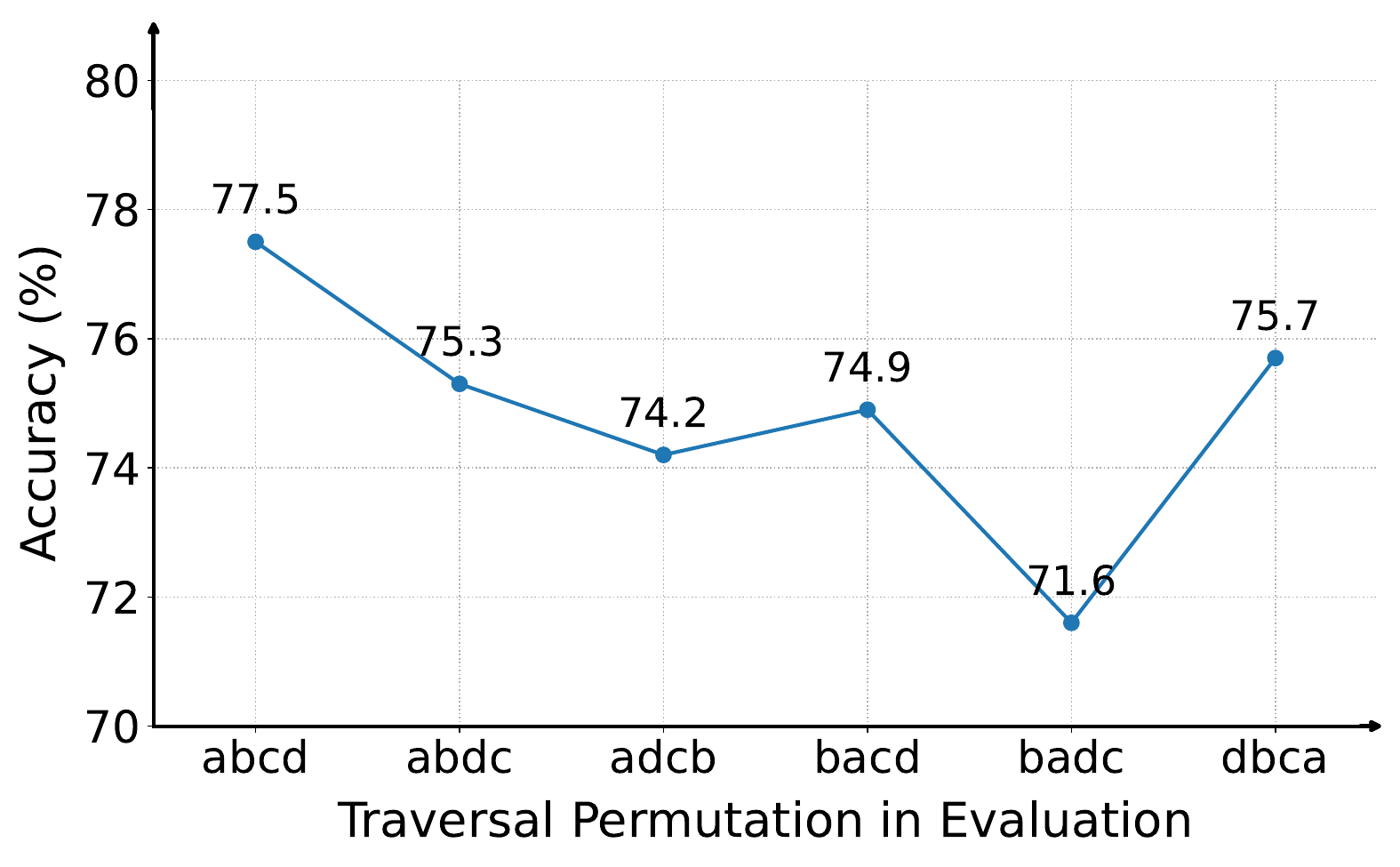}
    \vspace{-10pt}
    \caption{Impact of traversal permutation during evaluation on CIFAR10-C dataset.}
    \label{eval}
  \end{minipage}
\end{figure}

\begin{wrapfigure}{r}{0.45\textwidth}
  \centering
  \includegraphics[width=\linewidth]{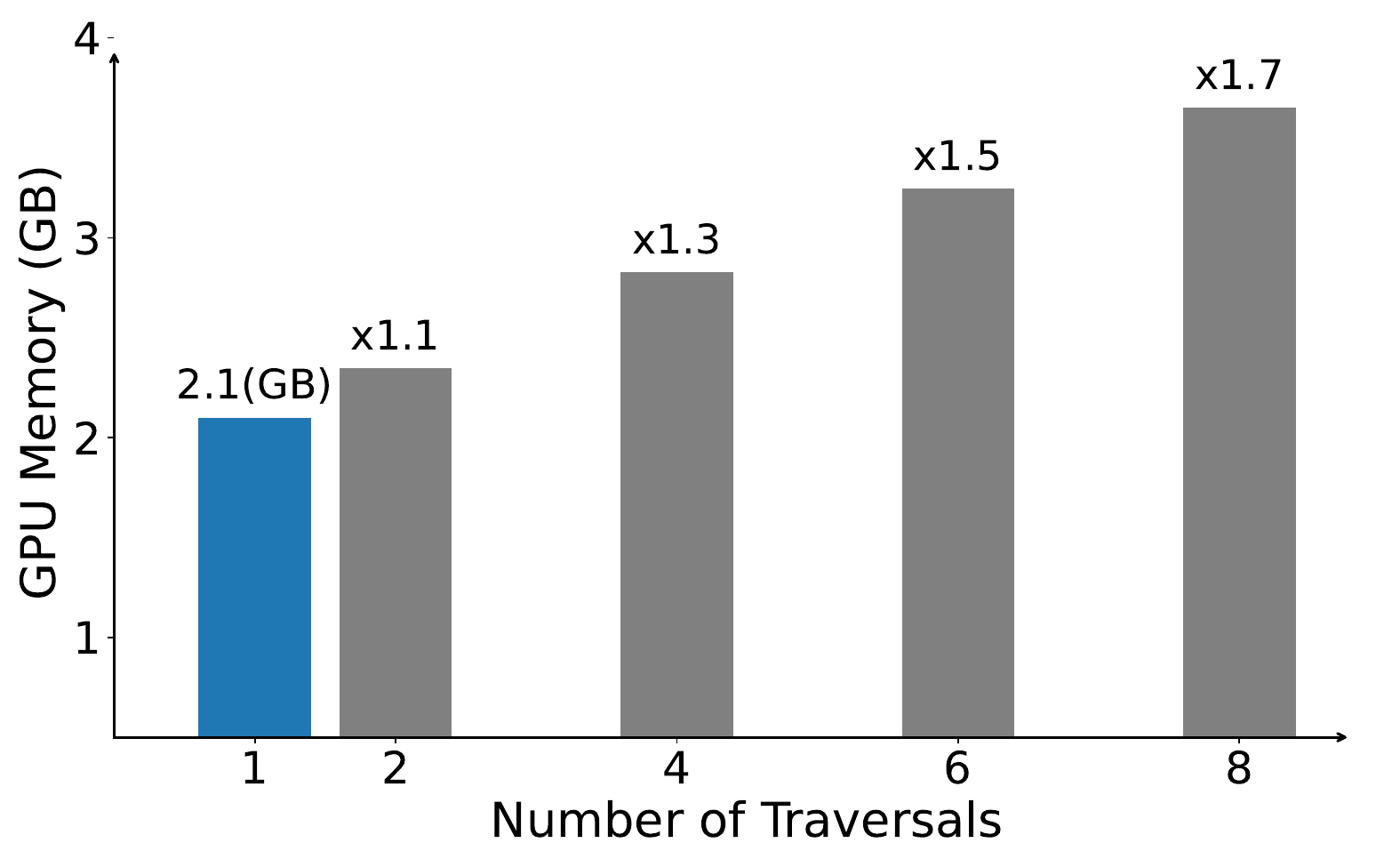}
  \vspace{-15pt}
  \caption{\small GPU memory usage across traversals.}
  \label{cc}
\end{wrapfigure}

\mypar{Aggregation Strategy.} We evaluate two baseline strategies for aggregation across traversal permutations: (1) an ensemble approach that averages model predictions from independently processed traversal orders, and (2) a repetition baseline, where the same traversal permutation is applied $k$ times, followed by weight averaging. As shown in Figure~\ref{aggregation}, both baselines yield lower performance on CIFAR10-C, with ensemble achieving 68.1\% and repetition 69.6\%, compared to 75.6\% with our method. 

The ensemble variant offers traversal diversity, but because its individual models are never weight-averaged, it lacks parameter alignment. The repetition baseline, in contrast, repeats the same traversal and updates the model sequentially; the weights, therefore, remain aligned, yet no traversal diversity is introduced. Both baselines perform worse than our method, demonstrating that neither unaligned traversal diversity (ensemble) nor aligned repetitions without diversity (repetition) is sufficient on its own. By combining these two ingredients, traversal diversity and parameter alignment, our approach achieves better generalization under distribution shift.

\mypar{Effect of Traversal Permutation in Evaluation.} Figure~\ref{eval} illustrates how accuracy varies with different traversal permutations during evaluation. We observe that the choice of permutation impacts the performance. The default permutation used in the original VMamba architecture, ``$abcd$'', achieves the highest accuracy at 77.5\%, while others, such as ``$badc$'', yield lower performance (e.g., 71.6\%). Given this sensitivity, we adopt the ``$abcd$'' traversal permutation, on which the model was trained, for all evaluations to ensure consistency and leverage the strongest baseline.

\paragraph{Computational Overhead.}
We evaluate GPU memory usage as a function of the number of traversal permutations used during parallel adaptation. Since only the \gls{ss2d} blocks are updated, we instantiate one \gls{ss2d} block per traversal while sharing the rest of the network. Traversals are batched and routed to their corresponding \gls{ss2d} blocks, and their outputs are then concatenated. This design minimizes computational overhead. As shown in Figure~\ref{cc}, memory usage increases moderately, 1.1$\times$, 1.3$\times$, 1.5$\times$, and 1.7$\times$ for 2, 4, 6, and 8 traversals, respectively, relative to the 2.1 GB baseline (measured with batch size 1). In sequential mode, forward and backward passes are performed separately for each traversal. We measure a per-traversal time of approximately 135 ms (with the batch size of 1), leading to a total cost of $K \times 135$ ms for $K$ traversals. As expected, memory usage remains nearly constant, while adaptation time grows linearly. Conversely, in parallel mode, adaptation time remains nearly constant, with moderate memory overhead, offering a practical trade-off given the performance gains. For more details, refer to the Supplementary Material.

\vspace{-10pt}
\section{Conclusion}
\vspace{-5pt}

In this paper, we introduced \gls{trust}, a test-time adaptation strategy specifically designed for vision-oriented state-space models. Our method addresses two key challenges in VMamba under distribution shift: (1) the strong inductive bias introduced by fixed traversal scans, and (2) the accumulation of domain-specific artifacts in hidden states during sequential processing. To overcome these issues, \gls{trust} exploits directional traversal mechanism of VMamba to generate complementary causal views of the input, performing adaptation based on each. The resulting models are aggregated via weighted parameter averaging, promoting convergence toward flatter, more robust regions in the loss landscape. Experiments on seven standard benchmarks show that \gls{trust} consistently outperform popular \gls{tta} models including Tent, SHOT, and SAR, in image classification tasks. Our method does not yet generalize to the medical domain. 

\vspace{-10pt}
\section{Limitations}
\vspace{-5pt}

\gls{trust} is designed for VMamba-style architectures and exploits the traversal orders available in the \gls{ss2d} module. Therefore, it is not directly applicable to \glspl{cnn} or standard \glspl{vit} without architectural modifications. This specificity is intentional, as our goal is to leverage Mamba’s sequential and linear-time computation under domain shift. Extending the core idea to other architectures remains an interesting direction for future work. Moreover, \gls{trust} introduces additional adaptation cost, as each selected traversal requires one forward and backward pass. In sequential mode, this cost scales linearly with the number of selected permutations $K$ (about 135 ms per traversal at batch size 1 in our setup). In practice, this overhead is controllable through $K$: our ablations show that performance saturates around $K{=}6$, providing a practical accuracy-efficiency trade-off. The choice of $K$ can therefore be adjusted based on the target architecture and latency budget. Finally, our theoretical analysis provides insight into why low-entropy traversal paths can lead to more stable recurrent dynamics and flatter loss regions, offering a principled explanation for the effectiveness of \gls{trust}. Extending this analysis to broader classes of domain shifts remains an interesting direction for future work.

\vspace{-10pt}
\section{Societal Impact}
\vspace{-5pt}

\gls{trust} aims to improve robustness under natural distribution shifts, but adaptive models should still be deployed carefully in real-world setting. Since \gls{trust} updates model parameters using pseudo-labels at test time, incorrect predictions may reinforce model errors under severe or unseen shifts, which is especially important in high-stakes applications such as medical or safety-critical decision making. While our experiments focus on natural corruptions and domain shifts, fairness, adversarial robustness, and energy-aware adaptation remain important directions for future evaluation before deployment in sensitive settings.
\section*{Acknowledgements}
\vspace{-5pt}

We appreciate the computational resources and support provided by Compute Canada and the Digital Research Alliance of Canada.
\bibliographystyle{unsrt}
\bibliography{main.bib} 
\newpage

\clearpage
\appendix            

\section*{\centering TRUST: Test-Time Refinement using Uncertainty-Guided~SSM~Traverses~-~Appendix}


\section{Pseudo-code} In this section, we give the pseudo-code for our proposed test-time adaptation method, \glsentryshort{trust}. This pseudo-code provides a concise summary of the key steps involved in our approach, offering a high-level abstraction of the implementation. Algorithm 1 outlines the overall \glsentryshort{trust} procedure for test-time adaptation. For each corruption in the evaluation set, the model is first reset to its original (pre-adaptation) weights. The input $x$ is then processed using the \textsc{Forward\_And\_Adapt} function (Algorithm~2), which outputs the adapted prediction and a list of model parameter sets $\theta$, each corresponding to a different traversal permutation used during adaptation. These parameters are averaged to obtain the final adapted parameters, $\theta_{\text{final}}$, which are loaded back into the model. The final prediction is then computed by re-evaluating the model on the same input $x$.

\begin{algorithm}[h]
\caption{\glsentryshort{trust}}
\begin{algorithmic}[1]
\For{each corruption}
    \State \Call{model.reset}{}
    \State $(\textit{out}, \theta) \gets$ \Call{model.forward\_and\_adapt}{$x$}
    \State $\theta_{\text{final}} \gets$ \Call{mean}{$\theta$}
    \State \Call{model.load\_state\_dict}{$\theta_{\text{final}}$}
    \State $\textit{out} \gets$ \Call{model.evaluate}{$x$}
\EndFor
\end{algorithmic}
\end{algorithm}

Algorithm 2 defines the \textsc{Forward\_And\_Adapt} function. For each traversal permutation $\pi_{i_k} \in \mathcal{P}$, the model performs a forward pass with input $x$ and permutation $\pi_{i_k}$. The cross-entropy loss between the model’s prediction and the pseudo-labels is computed, and a gradient descent step is performed to update the model parameters. Each updated parameter state is stored in the list $\theta$. After all permutations have been processed, the method returns the final output and the list of different model parameters.

\begin{algorithm}[h]
\caption{\textsc{Forward\_And\_Adapt}$(x)$}
\begin{algorithmic}[1]
\State $\theta \gets$ empty list
\For{each $\pi_{i_k} \in \mathcal{P}$}
    \State $\textit{out} \gets$ model($x, \pi_{i_k}$)
    \State $\textit{loss} \gets \text{CE}(\textit{out}, \textit{pseudo\_labels})$
    \State loss.backward()
    \State optimizer.step()
    \State optimizer.zero\_grad()
    \State $\theta.\text{append}(\text{model.parameters})$
\EndFor
\State \Return $(\textit{out}, \theta)$
\end{algorithmic}
\end{algorithm}

\section{Finetuning Process}
The backbone model, VMamba, was originally trained on ImageNet-C, which contains 1,000 classes. To enable test-time use on corrupted datasets with different label sets, we fine-tune only the classifier layer on the clean version of each target dataset, keeping the rest of the model frozen. This procedure is applied to CIFAR10-C, CIFAR100-C, and PACS, using a learning rate of 5e-4 for 300 epochs. For PACS, which includes four domains (photo, art painting, cartoon, and sketch), we follow the standard protocol: one domain is held out for evaluation while training on the remaining three. Specifically, we use the photo domain as the held-out test set. For datasets such as ImageNet-S, ImageNet-V2, and ImageNet-R, which share the same label space as ImageNet, no fine-tuning is required.

\section{Efficient Parallel Implementation}
\begin{figure}[h]
    \centering
    \includegraphics[trim=0 400 0 0, clip, width=0.7\textwidth]{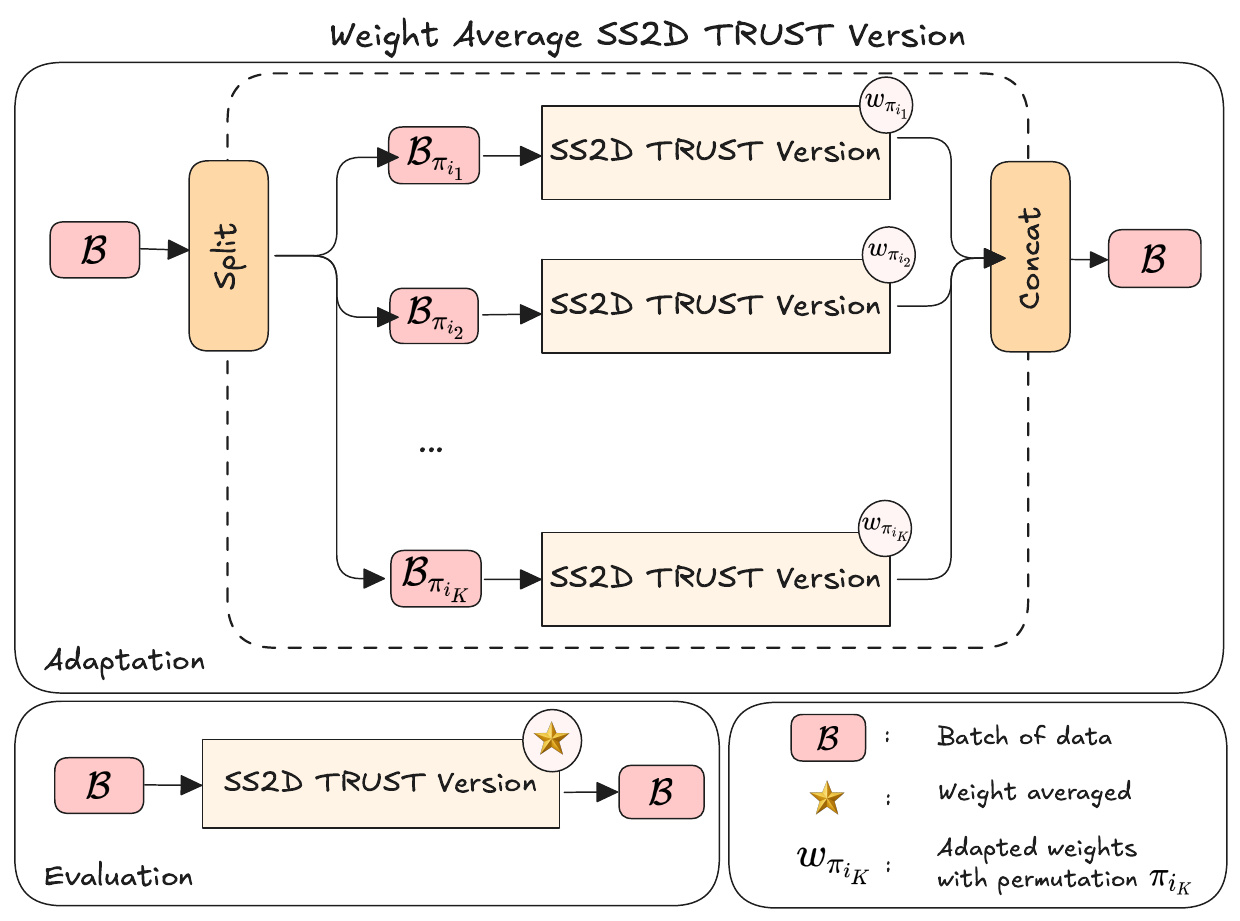}
    \vspace{-10pt}
    \caption{Detailed diagram of \glsentryshort{trust} in Parallel mode.}
    \label{main_figure_supp}
\end{figure}

We mentioned in the main paper (Computational Overhead section) that our method supports an efficient parallel adaptation strategy. Figure~\ref{main_figure_supp} provides a detailed diagram of this process. In parallel mode, we handle $K$ traversal permutations simultaneously. A batch $\mathcal{B}$ is first split into $K$ subsets, each corresponding to a different permutation $\pi_{i_k}$. Each subset is then passed to an independent \glsentryshort{ss2d} TRUST Version block, where the \glsentryshort{ss2d} parameters are adapted in parallel while the rest of the model remains shared across all paths, this design significantly reduces memory usage.

After adaptation, the outputs are concatenated back into a single batch, which allows for efficient GPU utilization. For evaluation, we perform a weight averaging step across all adapted \glsentryshort{ss2d} modules, producing a single unified \glsentryshort{ss2d} TRUST Version block. This averaged block is then used for inference on the full batch. By leveraging parallelism in both data and traversal space, our method achieves scalable and low-overhead test-time adaptation while preserving performance across permutations.

\section{Standard Mode} Table~\ref{domain_shift_standard} reports Top-1 accuracy under various corruption types at severity level 5 for CIFAR10-C, CIFAR100-C, and ImageNet-C in the standard setting, where the model is reset for each test batch. On CIFAR10-C, \glsentryshort{trust} achieves the highest mean accuracy of $66.2\%$, surpassing Tent and SHOT by $0.2\%$. On CIFAR100-C, it reaches $41.7\%$, outperforming all baselines by at least $0.4\%$, and improving over its naive variant by $0.5\%$, underscoring the benefits of permutation-based refinement. On the large-scale ImageNet-C, \glsentryshort{trust} attains $39.9\%$, exceeding Tent by $1.1\%$, and outperforming its naive counterpart by $1.2\%$. These results highlight the scalability and robustness of permutation-aware adaptation across diverse datasets. Table~\ref{domain_shift_standard} evaluates performance under domain-level distribution shifts. \glsentryshort{trust} achieves $31.6\%$ on \textbf{ImageNet-S}, $62.3\%$ on \textbf{ImageNet-V2}, and $31.5\%$ on \textbf{ImageNet-R}. On the \textbf{PACS} benchmark, it obtains $71.3\%$, outperforming its naive variant by $4.6\%$. While gains are narrower in these less corrupted settings, \glsentryshort{trust} consistently improves generalization, particularly under significant domain shifts.

\begin{table*}[h]
\setlength\tabcolsep{4pt}
\centering
\resizebox{0.9\textwidth}{!}{
\small
\begin{tabular}{l|ccccccc}
\toprule
\textbf{Method} & CIFAR10-C & CIFAR100-C & ImageNet-C & ImageNet-S & ImageNet-V2 & ImageNet-R & PACS \\
\midrule
Source only             & 65.9\same{0.0} & 41.2\same{0.0} & 38.7\same{0.0} & 31.4\same{0.0} & 62.2\same{0.0} & 31.3\same{0.0} & 66.7\same{0.0} \\
ETA                     & 65.8\worse{0.1} & 41.2\same{0.0} & 38.8\better{0.1} & 31.4\same{0.0} & 62.2\same{0.0} & 31.4\better{0.1} & 66.7\same{0.0} \\
LAME                    & 65.9\same{0.0} & 41.2\same{0.0} & 38.8\better{0.1} & 31.4\same{0.0} & 62.2\same{0.0} & 31.4\same{0.0} & 66.7\same{0.0} \\
SAR                     & 66.0\better{0.1} & 41.3\better{0.1} & 38.8\better{0.1} & 31.4\same{0.0} & 62.2\same{0.0} & 31.4\better{0.1} & 67.1\better{0.4} \\
SHOT                    & 66.0\better{0.1} & 41.3\better{0.1} & 38.9\better{0.2} & 31.4\same{0.0} & 62.2\same{0.0} & 31.4\better{0.1} & 67.3\better{0.6} \\
Tent                    & 66.0\better{0.1} & 41.3\better{0.1} & 38.8\better{0.1} & 31.4\same{0.0} & 62.2\same{0.0} & 31.4\better{0.1} & 67.2\better{0.5} \\                    
\glsentryshort{trust} naive            & 65.9\same{0.0} & 41.2\same{0.0} & 38.9\better{0.2} & 31.5\better{0.1} & 62.3\better{0.1} & 31.5\better{0.2} & 66.8\better{0.1} \\
\ccol \glsentryshort{trust}            & \ccol \textbf{66.2\better{0.3}} & \ccol \textbf{41.7\better{0.5}} & \ccol \textbf{39.9\better{1.2}} & \ccol \textbf{31.6\better{0.2}} & \ccol \textbf{62.3\better{0.1}} & \ccol \textbf{31.5\better{0.2}} & \ccol \textbf{71.3\better{4.6}} \\
\bottomrule
\end{tabular}
}
\caption{Top-1 classification accuracy (\%) across datasets in standard setting. For CIFAR10-C, CIFAR100-C, and ImageNet-C, values are averaged over all corruptions; for ImageNet-S, V2, R, and PACS, they reflect test set accuracy. Increases/decreases in mean accuracy compared to performing no adaptation (Source only) is highlighted in green/red color.}
\label{domain_shift_standard}
\end{table*}

\section{\glsentryshort{trust} with Batch Norm Adaptation} In this experiment, we compare two adaptation strategies within our framework: updating only the \glsentryshort{bn} layers versus updating the \glsentryshort{ss2d} parameters. As shown in Table~\ref{bn}, our method outperforms Tent in both settings, by 3.1\% when using BatchNorm adaptation and by 11.8\% with SS2D adaptation. We opt to proceed with SS2D adaptation, as traversal permutations directly influence Mamba-specific parameters encoded in the SS2D blocks. Updating these parameters is therefore essential to fully capture the effects of traversal-based modifications.

\begin{table*}[h]
\setlength\tabcolsep{4pt}
\centering
\resizebox{0.9\textwidth}{!}{
\small
\begin{tabular}{l|l|ccccccccccccccc|c}
\toprule
& Method & \rotatebox{90}{gaussian} & \rotatebox{90}{shot} & \rotatebox{90}{impulse} & \rotatebox{90}{defocus} & \rotatebox{90}{glass} & \rotatebox{90}{motion} & \rotatebox{90}{zoom} & \rotatebox{90}{frost} & \rotatebox{90}{snow} & \rotatebox{90}{fog} & \rotatebox{90}{brightness} & \rotatebox{90}{contrast} & \rotatebox{90}{elastic} & \rotatebox{90}{pixelate} & \rotatebox{90}{jpeg} & Mean \\
\cmidrule{2-18}
& Source only                & 24.3 & 26.1 & 25.1 & 22.2 & 23.2 & 35.4 & 43.2 & 49.3 & 48.4 & 56.9 & 70.0 & 26.8 & 45.1 & 43.7 & 41.4 & 38.7\same{0.0} \\
\midrule
\multirow{2}{*}{\centering\arraybackslash\parbox{0.7cm}{\centering \glsentryshort{bn}}}
& Tent                       & 27.8 & 30.0 & 28.8 & 24.9 & 25.9 & 38.0 & 45.5 & 51.0 & 51.3 & 59.1 & 70.6 & 30.0 & 48.2 & 47.8 & 45.7 & 41.7\same{0.0} \\
& \ccol \glsentryshort{trust}      & \ccol 32.8 & \ccol 35.1 & \ccol 34.0 & \ccol 26.8 & \ccol 28.5 & \ccol 40.8 & \ccol 47.7 & \ccol 52.9 & \ccol 53.8 & \ccol 61.1 & \ccol 71.3 & \ccol 34.1 & \ccol 50.7 & \ccol 51.8 & \ccol 50.2 & \ccol 44.8\better{3.1}  \\
\midrule
\multirow{2}{*}{\centering\arraybackslash\parbox{0.7cm}{\centering SS2D}}
& Tent                       & 29.6 & 31.7 & 34.7 & 25.1 & 22.0 & 45.7 & 44.8 & 46.0 & 55.8 & 62.2 & 69.8 & 32.5 & 52.8 & 56.8 & 54.3 & 44.3\same{0.0} \\
& \ccol \glsentryshort{trust}      & \ccol 46.8 & \ccol 49.4 & \ccol 48.5 & \ccol 42.8 & \ccol 40.8 & \ccol 57.1 & \ccol 57.9 & \ccol 57.3 & \ccol 61.7 & \ccol 66.8 & \ccol 71.9 & \ccol 54.9 & \ccol 61.4 & \ccol 63.6 & \ccol 60.2 & \ccol 56.1\better{11.8} \\
\bottomrule
\end{tabular}
}
\caption{Comparison of adaptation strategies using \glsentryshort{bn} and \glsentryshort{ss2d} parameters on ImageNet-C dataset. Increases/decreases in mean accuracy compared to Tent is highlighted in green/red color.}
\label{bn}
\end{table*}

\section{Mean Entropy of Different Traversal Permutations} 
\begin{figure}[h]
    \centering
    \includegraphics[width=\linewidth]{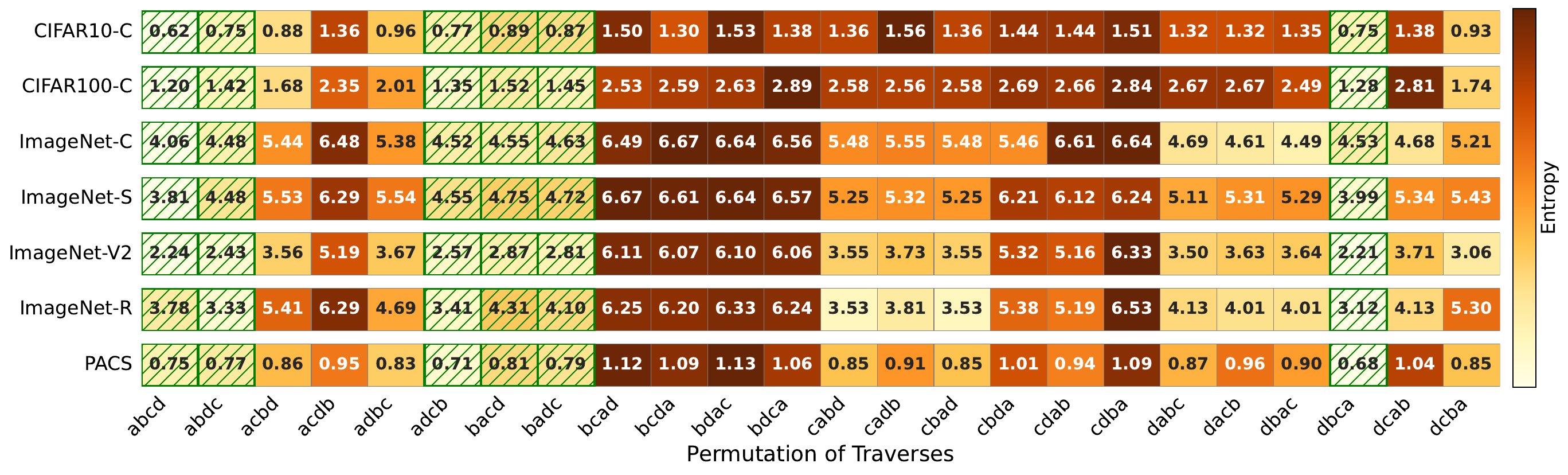}
    \vspace{-15pt}
    \caption{Mean entropy of different traversal permutation across seven benchmarks.}
    \label{heatmap_combined}
\end{figure}

As illustrated in Figure~\ref{heatmap_combined}, varying the order of spatial traversals in VMamba leads to substantial differences in mean entropy across datasets, revealing the sensitivity of the model’s internal representations to traversal patterns. Higher entropy values indicate less confident predictions, often aligning with reduced robustness under distributional shifts. This highlights the critical role of traversal selection in ensuring reliable adaptation. To offer a comprehensive analysis, we report entropy values for all permutations across CIFAR10-C, CIFAR100-C, ImageNet-C, ImageNet-S, ImageNet-V2, ImageNet-R, and PACS. Notably, the top-2, top-4, and top-6 traversal permutations with the lowest entropy, highlighted with green cross-hatching (i.e., diagonal lines overlaid on the heatmap cells), consistently reappear across datasets. This pattern suggests that certain traversal orders inherently yield more stable and confident model outputs, providing a principled foundation for selecting effective traversal subsets in our approach.

\section{Combination of \glsentryshort{trust} with Augmentation} According to Figure 4 in the main paper, Jitter augmentation yields the highest performance among the augmentation strategies evaluated. Building on this, we examine the effectiveness of combining Jitter with our proposed method, \glsentryshort{trust}, under two augmentation settings: (1) applying different Jitter augmentations for each traversal permutation, and (2) applying different Jitter augmentations per batch. The former setting achieves a modest improvement of $0.3\%$ over the source-only baseline, while the latter results in a slightly higher gain of $1.8\%$. However, both improvements are substantially lower than the performance boost achieved by the original \glsentryshort{trust}, which yields an $11.6\%$ increase. These results highlights that although augmentation contributes to performance, the core strength of \glsentryshort{trust} lies in its ability to adapt representations based on traversal-specific dynamics rather than augmentation diversity alone.

\begin{table*}[h]
\setlength\tabcolsep{4pt}
\centering
\resizebox{0.95\textwidth}{!}{
\small
\begin{tabular}{l|ccccccccccccccc|c}
\toprule
Method & \rotatebox{90}{gaussian} & \rotatebox{90}{shot} & \rotatebox{90}{impulse} & \rotatebox{90}{defocus} & \rotatebox{90}{glass} & \rotatebox{90}{motion} & \rotatebox{90}{zoom} & \rotatebox{90}{frost} & \rotatebox{90}{snow} & \rotatebox{90}{fog} & \rotatebox{90}{brightness} & \rotatebox{90}{contrast} & \rotatebox{90}{elastic} & \rotatebox{90}{pixelate} & \rotatebox{90}{jpeg} & Mean \\
\midrule
Source only                            & 46.8 & 48.4 & 45.0 & 73.5 & 52.6 & 73.0 & 78.7 & 71.8 & 75.8 & 77.3 & 85.7 & 69.6 & 63.7 & 67.9 & 59.0 & 65.9\same{0.0} \\
\glsentryshort{trust} + Jitter per permutation  & 53.6 & 54.2 & 54.0 & 63.8 & 48.0 & 69.3 & 77.1 & 75.4 & 75.2 & 77.2 & 85.8 & 76.6 & 59.1 & 66.5 & 57.8 & 66.2\better{0.3} \\
\glsentryshort{trust} + Jitter per batch & 57.0 & 53.4 & 57.2 & 70.0 & 48.9 & 65.6 & 79.6 & 77.5 & 77.9 & 77.3 & 88.2 & 79.3 & 57.9 & 65.4 & 60.6 & 67.7\better{1.8} \\
\ccol \glsentryshort{trust}                 & \ccol 63.1 & \ccol 67.8 & \ccol 70.3 & \ccol 81.0 & \ccol 64.5 & \ccol 81.4 & \ccol 85.0 & \ccol 83.2 & \ccol 85.4 & \ccol 85.8 & \ccol 90.1 & \ccol 85.7 & \ccol 72.1 & \ccol 79.1 & \ccol 68.6 & \ccol 77.5\better{11.6} \\
\bottomrule
\end{tabular}
}
\caption{Comparison of augmentation impact on CIFAR10-C dataset. Increases/decreases in mean accuracy compared to performing no adaptation (Source only) is highlighted in green/red color.}
\label{aug_combination}
\end{table*}

\section{High Entropy Traversal Permutations} We have also tested the performance of \glsentryshort{trust} using the top-\textit{k} high-entropy (i.e., less confident) traversal permutations. As shown in Table~\ref{high_entropy_traversal}, this setting leads to a significant performance drop of $12.5\%$ compared to the source-only baseline. This highlights the critical role of traversal selection in our method. High-entropy permutations, which correspond to uncertain and unstable model predictions, introduce noise into the adaptation process and hinder effective generalization. These findings further support our strategy of entropy-based traversal filtering, where low-entropy permutations are prioritized to ensure reliable and robust adaptation.

\begin{table*}[h]
\setlength\tabcolsep{4pt}
\centering
\resizebox{0.95\textwidth}{!}{
\small
\begin{tabular}{l|ccccccccccccccc|c}
\toprule
Method & \rotatebox{90}{gaussian} & \rotatebox{90}{shot} & \rotatebox{90}{impulse} & \rotatebox{90}{defocus} & \rotatebox{90}{glass} & \rotatebox{90}{motion} & \rotatebox{90}{zoom} & \rotatebox{90}{frost} & \rotatebox{90}{snow} & \rotatebox{90}{fog} & \rotatebox{90}{brightness} & \rotatebox{90}{contrast} & \rotatebox{90}{elastic} & \rotatebox{90}{pixelate} & \rotatebox{90}{jpeg} & Mean \\
\midrule
Source only                                 & 46.8 & 48.4 & 45.0 & 73.5 & 52.6 & 73.0 & 78.7 & 71.8 & 75.8 & 77.3 & 85.7 & 69.6 & 63.7 & 67.9 & 59.0 & 65.9\same{00.0} \\
\glsentryshort{trust} (top-k high entropy)  & 38.5 & 32.2 & 36.4 & 57.9 & 37.7 & 61.5 & 64.9 & 52.5 & 64.0 & 70.3 & 74.1 & 49.3 & 57.9 & 56.2 & 48.2 & 53.4\worse{12.5} \\
\ccol \glsentryshort{trust} (top-k low entropy)   & \ccol 63.1 & \ccol 67.8 & \ccol 70.3 & \ccol 81.0 & \ccol 64.5 & \ccol 81.4 & \ccol 85.0 & \ccol 83.2 & \ccol 85.4 & \ccol 85.8 & \ccol 90.1 & \ccol 85.7 & \ccol 72.1 & \ccol 79.1 & \ccol 68.6 & \ccol 77.5\better{11.6} \\
\bottomrule
\end{tabular}
}
\caption{Comparison of \glsentryshort{trust} performance with low and high entropy traversal permutations on CIFAR10-C dataset.}
\label{high_entropy_traversal}
\end{table*}


\section{\gls{trust} Application beyond Classification Settings}
We conducted additional experiments on segmentation tasks using various datasets, including Pascal VOC21 \cite{v21} and Pascal Context 59 \cite{p59}, applying corruptions following the protocol of \cite{noori2025test}. The results demonstrate that \gls{trust} performs well in segmentation, outperforming methods like Tent. This further supports the generalizability and effectiveness of our approach beyond classification settings.

\begin{table*}[htbp]
\setlength\tabcolsep{4pt}
\centering
\resizebox{0.95\textwidth}{!}{
\small
\begin{tabular}{l|l|ccccccccccccccc|c}
\toprule
\rotatebox{90}{Dataset} & Method & \rotatebox{90}{gaussian noise} & \rotatebox{90}{shot noise} & \rotatebox{90}{impulse noise} & \rotatebox{90}{defocus blur} & \rotatebox{90}{glass blur} & \rotatebox{90}{motion blur} & \rotatebox{90}{zoom blur} & \rotatebox{90}{frost} & \rotatebox{90}{snow} & \rotatebox{90}{fog} & \rotatebox{90}{brightness} & \rotatebox{90}{contrast} & \rotatebox{90}{elastic} & \rotatebox{90}{pixelate} & \rotatebox{90}{jpeg compression} & Mean \\
\midrule
\parbox[t]{4mm}{\multirow{3}{*}{\rotatebox[origin=c]{90}{V21}}} 
& Source only & 29.1 & 33.1 & 28.3 & 21.0 & 8.2  & 33.1 & 25.4 & 50.9 & 50.3 & 70.7 & 76.5 & 63.9 & 25.5 & 22.2 & 59.2 & 39.8\same{00.0} \\
& Tent        & 33.0 & 35.7 & 32.0 & 22.3 & 14.7 & 38.2 & 25.3 & 46.5 & 49.0 & 60.2 & 63.9 & 66.2 & 38.5 & 28.8 & 43.9 & 39.9\same{00.0} \\
& \ccol \gls{trust} & \ccol \textbf{38.8} & \ccol \textbf{42.0} & \ccol \textbf{38.7} & \ccol \textbf{29.8} & \ccol \textbf{22.6} & \ccol \textbf{45.1} & \ccol \textbf{29.8} & \ccol \textbf{50.5} & \ccol \textbf{53.5} & \ccol \textbf{63.4} & \ccol \textbf{66.4} & \ccol \textbf{68.5} & \ccol \textbf{45.1} & \ccol \textbf{37.7} & \ccol \textbf{48.6} & \ccol \textbf{45.4} \better{5.6} \\
\midrule
\parbox[t]{4mm}{\multirow{3}{*}{\rotatebox[origin=c]{90}{P59}}} 
& Source only & 17.1 & 19.6 & 17.4 & 27.4 & 14.9 & 29.2 & 19.5 & 30.2 & 28.5 & 42.1 & 50.8 & 41.0 & 23.9 & 30.4 & 38.4 & 28.7\same{00.0} \\
& Tent        & 17.6 & 18.9 & 17.8 & 22.2 & 15.9 & 27.5 & 17.9 & 26.9 & 30.0 & 36.7 & 41.9 & 42.9 & 25.8 & 28.2 & 28.3 & 26.6\same{00.0} \\
& \ccol \gls{trust} & \ccol \textbf{24.4} & \ccol \textbf{27.4} & \ccol \textbf{25.4} & \ccol \textbf{24.6} & \ccol \textbf{21.2} & \ccol \textbf{30.1} & \ccol \textbf{19.8} & \ccol \textbf{29.8} & \ccol \textbf{32.8} & \ccol \textbf{39.2} & \ccol \textbf{42.4} & \ccol \textbf{43.2} & \ccol \textbf{31.5} & \ccol \textbf{36.1} & \ccol \textbf{31.6} & \ccol \textbf{30.6} \better{1.9}\\
\bottomrule
\end{tabular}
}
\caption{Segmentation performance under different corruptions of V21 and P59 datasets.}
\label{tab:segmentation_v21_p59}
\end{table*}

\section{Performance Comparison with ViT-based Method}
To provide a fair evaluation, we adapted the ViT-based method from \cite{foa} by replacing its backbone with VMamba and modifying it to be compatible with VMamba backbone. Specifically, we replaced the CLS token (absent in VMamba) with the mean of all tokens and adjusted the number of learnable tokens to match VMamba’s dimensionality. This ensures that both methods operate under the same architectural constraints. Our experimental results demonstrate that TRUST, which is specifically designed to exploit VMamba’s traversal mechanism, significantly outperforms the adapted baseline, highlighting its robustness under distribution shift.

\begin{table*}[h]
\setlength\tabcolsep{4pt}
\centering
\resizebox{0.95\textwidth}{!}{
\small
\begin{tabular}{l|ccccccccccccccc|c}
\toprule
Method & \rotatebox{90}{gaussian} & \rotatebox{90}{shot} & \rotatebox{90}{impulse} & \rotatebox{90}{defocus} & \rotatebox{90}{glass} & \rotatebox{90}{motion} & \rotatebox{90}{zoom} & \rotatebox{90}{frost} & \rotatebox{90}{snow} & \rotatebox{90}{fog} & \rotatebox{90}{brightness} & \rotatebox{90}{contrast} & \rotatebox{90}{elastic} & \rotatebox{90}{pixelate} & \rotatebox{90}{jpeg} & Mean \\
\midrule
Source only & 24.3 & 26.1 & 25.1 & 22.2 & 23.2 & 35.4 & 43.2 & 49.3 & 48.4 & 56.9 & 70.0 & 26.8 & 45.1 & 43.7 & 41.4 & 38.7 \same{00.0} \\
Tent        & 27.8 & 30.0 & 28.8 & 24.9 & 25.9 & 38.0 & 45.5 & 51.0 & 51.3 & 59.1 & 70.6 & 30.0 & 48.2 & 47.8 & 45.7 & 41.7 \same{00.0} \\
FOA         & 17.8 & 19.4 & 18.5 & 15.1 & 18.2 & 22.7 & 28.6 & 38.7 & 33.9 & 44.3 & 57.4 & 22.7 & 38.2 & 40.9 & 41.7 & 30.5 \worse{8.2} \\
\ccol TRUST       & \ccol 46.8 & \ccol 49.4 & \ccol 48.5 & \ccol 42.8 & \ccol 40.8 & \ccol 57.1 & \ccol 57.9 & \ccol 57.3 & \ccol 61.7 & \ccol 66.8 & \ccol 71.9 & \ccol 54.9 & \ccol 61.4 & \ccol 63.6 & \ccol 60.2 & \ccol 56.1 \better{17.4} \\
\bottomrule
\end{tabular}
}
\caption{Performance comparison across corruption types with the ViT-based Method.}
\label{tab:corruption_results}
\end{table*}

\section{Weight Diversity Across Traversals}
To analyze the consistency of the SS2D block parameters under different traversal permutations, we compute statistics based on the L2 norms of each parameter tensor across six traversal-adapted models. We extract each weight or bias tensor from each model, flatten it, and compute its L2 norm. We then calculate the mean of these norms to reflect the overall magnitude across traversals. To quantify variability, we also compute the standard deviation across the corresponding elements of these tensors and report the average of these deviations.

\begin{figure}[h]
    \centering
    \includegraphics[width=.8\linewidth]{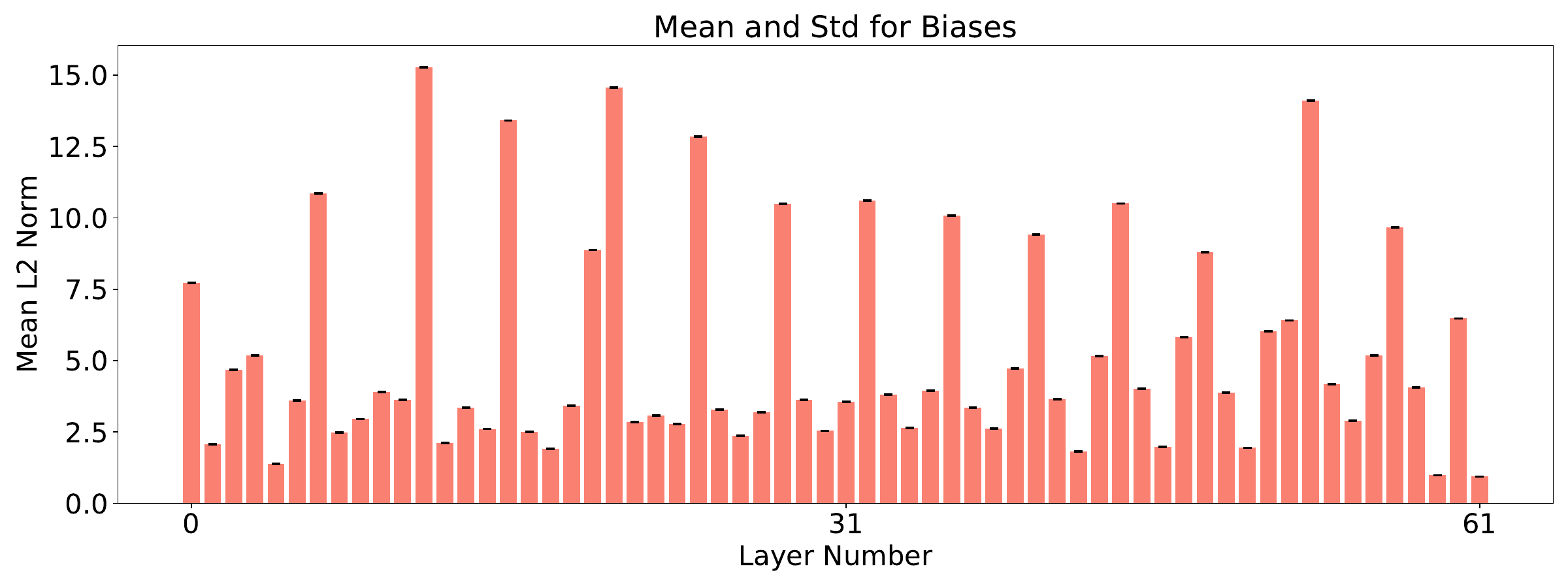}
    \vspace{-10pt}
    \caption{Mean and standard deviation of the L2 norm for the bias parameters}
    \label{biases}
\end{figure}

\begin{figure}[h]
    \centering
    \includegraphics[width=.8\linewidth]{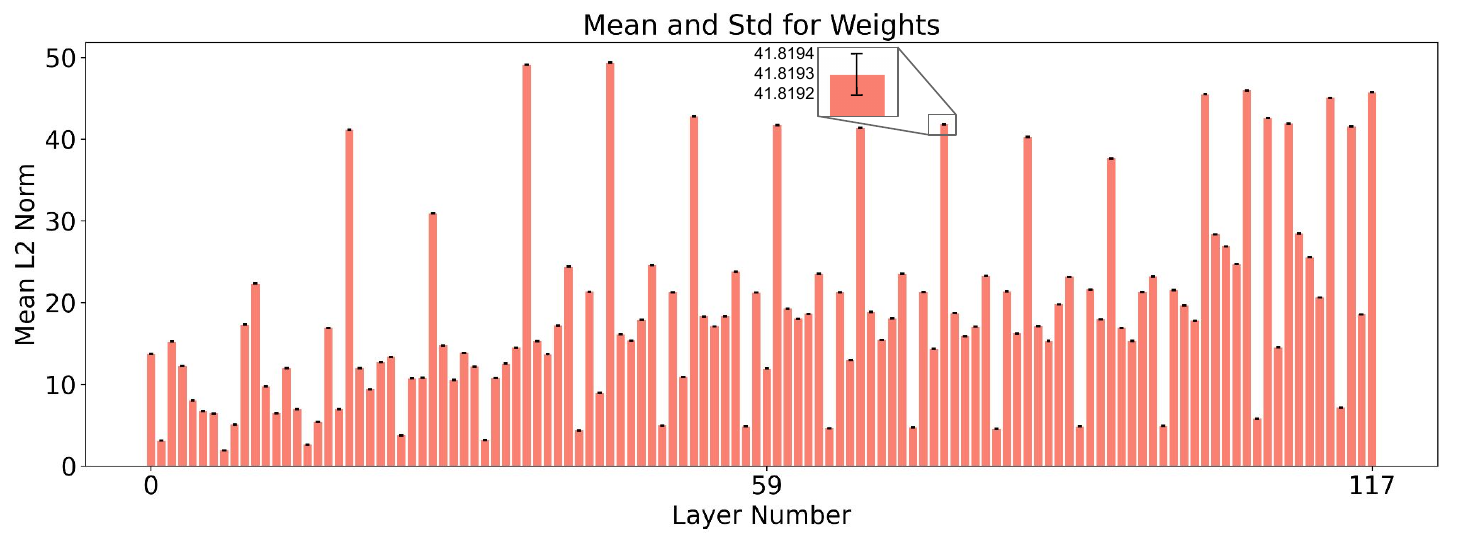}
    \vspace{-10pt}
    \caption{Mean and standard deviation of the L2 norm for the weight parameters}
    \label{weights}
\end{figure}

Figure~\ref{biases} shows the mean and standard deviation of the L2 norm for the bias parameters, while Figure~\ref{weights} shows the same for the weight parameters. These visualizations indicate that although the magnitude of parameters varies across layers, the variation across traversals remains relatively low. This suggests that SS2D block parameters adapted with different traversal orders remain geometrically close in parameter space. Such consistency is conceptually aligned with the findings in the Model Stock method~\cite{jang2024model}, which emphasizes the benefits of maintaining proximity in the weight space across fine-tuned or adapted models, such as enabling effective weight averaging and improving generalization. This experiment is conducted using the CIFAR-10C dataset.

\section{Sensitivity to Traversal Order}
\glspl{ssm} are inherently direction-sensitive, meaning that the order in which input patches are traversed has a significant impact on model performance. According to the Mamba formulation Eq.~\ref{mamba}, the hidden state at each step $h(t-1)$ directly influences the next $h(t)$. Therefore, the choice of traversal order propagates its effects through the entire sequence of hidden-state updates. This directional dependency also comes from the fact that the hidden states learned during training are aligned with the original traversal order. When a different patch order is used at test time, the model is exposed to a sequence it has not encountered before, resulting in a mismatch between learned and test-time dynamics. As noted in the Mamba paper, this directional bias can cause the model to encode information more effectively along the original traversal path.

\section{Offline Phase Scalability}
The scalability of the offline selection phase is mainly determined by the number of traversal permutations. Each permutation requires only a forward pass through the model, with no backpropagation or parameter updates. As a result, the computational cost grows linearly with the number of candidate permutations. Memory usage, however, does not grow with the number of permutations, since the same model is reused for each forward pass. The required memory is therefore equivalent to running a single model evaluation. This makes the permutation search computationally predictable, while keeping the memory footprint fixed.

\section{Theoretical Analysis of Confidence-Based Traversal Paths}

We provide a deeper theoretical justification of why confidence-based paths improve generalization based on SWAD \cite{swad}. In our setting, each traversal permutation $\pi_i$ defines a distinct causal ordering through which VMamba processes input patches, resulting in different trajectories of hidden states $h(t)$. Low-entropy (high-confidence) permutations consistently correspond to stable recurrent dynamics, where the influence of corruptions is minimized. This stabilizes the hidden-state evolution, yielding low-variance predictions and reducing sensitivity to domain-specific perturbations. From a loss landscape perspective, such paths correspond to regions of low curvature, as smoother predictions typically reflect flatter neighborhoods around $\theta_i$ (parameters of adapted model with $\pi_i$).

Therefore, selecting low-entropy permutations acts as a proxy for identifying solutions that are both dynamically stable (in terms of hidden-state recurrence) and geometrically robust (in terms of curvature). Averaging over the adapted weights from these confidence paths further concentrates the model in a flat, low-loss region of parameter space. This synergy between entropy-based selection and weight averaging aligns directly with the generalization theory of SWAD \cite{swad}, and provides a principled explanation for the observed robustness of \gls{trust} across domains.

\section{TRUST Latency}

Our method consists of three stages: Offline phase: We evaluate our pre-trained model under different traversal permutations and select the top-k permutations with the lowest entropy values, corresponding to the most confident predictions. Since no model adaptation occurs at this stage, the procedure is equivalent to a forward-pass evaluation and is computationally lightweight. For instance, on the PACS dataset, evaluating for a single permutation takes only 4 seconds for all images in this dataset. Extending this across all 24 permutations results in a total evaluation time of 1.5 minutes, which is a one-time cost. Our experiments show that the top-K permutations remain consistent across all test datasets, as visualized in Figure~\ref{heatmap_combined} of the supplementary material. This consistency stems from the similarity of these permutations to the original traversal order used during pre-training, as reflected in their high confidence scores. Therefore, this procedure can be performed only once on a single dataset to determine the appropriate Ks, and then used across all test datasets. Adaptation stage: We measured the latency for the batch size of 128 used in our experiments for forward and backward passes, which results in approximately 1.2 second per permutation. Evaluation stage: In this phase, we first perform weight averaging, which involves a simple averaging of the model weights across selected permutations (0.018 second). The subsequent evaluation functions similarly to the offline stage, no adaptation is performed, and we simply test the model (0.2 second). Our overall latency per permutation is: $1.2 + 0.018 + 0.2 = 1.41$ second.

\section{Comparison with Single-Pass Augmentation-Free TTA Methods}
Recent weight-averaging methods provide a single-pass, augmentation-free alternatives for \gls{tta}. For example, DPCore~\cite{zhang2024dpcore} averages prompt pools for the same input image, reducing the need for augmentation-based evaluation. However, DPCore does not perform model weight averaging. Instead, it constructs a weighted prompt by interpolating prompt tokens from a learned coreset, where the weights are computed based on the distance between the current batch’s feature statistics (extracted without prompts) and those stored in the coreset. As described in DPCore: “each batch requires two additional forward passes (one without prompt, one with weighted prompt)”. Therefore, while DPCore avoids per-image multi-prompt trials, it is strictly dual-pass. Moreover, DPCore learns a new prompt from scratch over 50 steps, and refines the existing prompt for 1 step. By contrast, TRUST adapts with a single-pass per selected traversal and the evaluation is single-pass after we perform weight averaging across traversals. It is worth mentioning that our method is not limited to K=6 traversal permutations; similar to DPCore dual-pass setup, it can benefit from as few as two traversal permutations (Table~\ref{tab:corruption_results}).

\begin{table}[h]
\centering
\small
\caption{Top-1 classification accuracy (\%) across datasets.}
\label{tab:corruption_results}
\begin{tabular}{lccc}
\toprule
Methods & CIFAR10-C (\%) & CIFAR100-C (\%) & ImageNet-C (\%) \\
\midrule
Source Only & 65.9 & 41.2 & 38.7 \\
Tent & 66.5 & 41.8 & 41.7 \\
\ccol TRUST (K=2) & \ccol 75.6 & \ccol 51.7 & \ccol 54.4 \\
\bottomrule
\end{tabular}
\end{table}

As DPCore and \gls{trust} use different architectures, a direct comparison of performance is not possible. Instead, we conducted a comparative evaluation using Tent as baseline. We report the relative latency of our method against DPCore and Tent. As stated in the DPCore paper, their method incurs approximately 1.8× the latency of Tent. In our case, with K = 2 traversals, our method incurs roughly 3.2× Tent’s latency (see Table~\ref{tab:runtime_comparison}). However, this increase is modest relative to the performance gains over Tent. Notably, TRUST achieves an additional 1.6\% improvement compared to DPCore.

\begin{table}[h]
\centering
\caption{Runtime comparison relative to Tent.}
\label{tab:runtime_comparison}
\begin{tabular}{lccc}
\toprule
Methods & Tent & DPCore & TRUST \\
\midrule
Time & 1.0 & 1.8$\times$ & 3.2$\times$ \\
\bottomrule
\end{tabular}
\end{table}

Furthermore, DPCore use source images during adaptation and depends on pre-computed source/reference statistics (typically from ~300 unlabeled source images, or from a proxy public dataset) to anchor its alignment objective. In contrast, TRUST is source- and reference-free: we do not require source images, source statistics, or proxy datasets at any stage.

Finally, TRUST and methods like DPCore operate in different architectural regimes: DPCore is prompt-centric and well-suited for ViTs, whereas TRUST is traversal-centric and tailored for Mamba-based models. Our aim is not to replace single-pass methods, but to offer an architecture-aligned TTA strategy that is both robust and efficient for state-space models.





\end{document}